\documentclass[12pt]{article}
\usepackage{verbatim,color}
\usepackage{graphicx}
\usepackage{setspace}
\usepackage{array}
\usepackage{epstopdf}
\usepackage{amsmath}
\usepackage{multirow}
\usepackage{caption}
\usepackage{enumitem}
\usepackage{algpseudocode}
\usepackage{array}
\usepackage{float}
\usepackage{caption}
\usepackage{subfigure}
\usepackage{geometry}
\usepackage{csquotes}
\usepackage{geometry}
\usepackage{sectsty}
\usepackage{bm}
\usepackage{rotating}
\usepackage{natbib, hhline}
\usepackage[linesnumbered,boxed,ruled]{algorithm2e}

\newcommand{\PreserveBackslash}[1]{\let\temp=\\#1\let\\=\temp}
\newcolumntype{C}[1]{>{\PreserveBackslash\centering}p{#1}}
\newcolumntype{R}[1]{>{\PreserveBackslash\raggedleft}p{#1}}
\newcolumntype{L}[1]{>{\PreserveBackslash\raggedright}p{#1}}

\newtheorem{remark}{Remark}
\usepackage{algorithm}

\begin{document}

		\begin{center}
			{\Large{\bf JST-RR Model: Joint Modeling of Ratings and Reviews in Sentiment-Topic Prediction}} \\
			\vskip 5pt
            Qiao Liang$^{\dag}$, Shyam Ranganathan$^{\ddag}$, Kaibo Wang$^{\dag}$, and Xinwei Deng$^{\ddag}$\footnote{Address for correspondence: Xinwei Deng, Associate Professor, Department of Statistics,
			Virginia Tech, Blacksburg, VA 24061 (E-mail: xdeng@vt.edu).} \\
			$^{\dag}$ Department of Industrial Engineering, Tsinghua University \\
			$^{\ddag}$ Department of Statistics, Virginia Tech
		\end{center}
		\vskip 2mm	
		
\begin{abstract}
Analysis of online reviews has attracted great attention with broad applications.
Often times, the textual reviews are coupled with the numerical ratings in the data.
In this work, we propose a probabilistic model to accommodate both textual reviews and overall ratings with consideration of their intrinsic connection for a joint sentiment-topic prediction.
The key of the proposed method is to develop a unified generative model where the topic modeling is constructed based on review texts
and the sentiment prediction is obtained by combining review texts and overall ratings.
The inference of model parameters are obtained by an efficient Gibbs sampling procedure.
The proposed method can enhance the prediction accuracy of review data and achieve an effective detection of interpretable topics and sentiments.
The merits of the proposed method are elaborated by the case study from Amazon datasets and simulation studies.

\end{abstract}

\newpage
\section{Introduction}

In modern service applications, there are increasing amounts of online reviews generated by users in recent years. The online reviews often contain both the text reviews and overall ratings. For example, the reviews in Amazon.com contain a review text on customer opinions of products or services, as well as the overall rating score on the general evaluation. Clearly, these user-generated contents can provide valuable information for both customers and online merchants \citep{liu2012sentiment}.
Among various research works on analyzing such review data,
topic identification \citep{titov2008modeling, blei2012probabilistic, airoldi2016improving} and sentiment classification \citep{bai2011predicting, taddy2013measuring, Ana2017} are two major directions.
The former aims to extract representing features or aspects of interest from discrete review words,
and the latter is to predict the semantic orientation of a review text.
With consideration of the inherent dependency between sentiment polarities and topics,
a simultaneous detection of correlated topics and sentiments serves as a critical function in the information retrieval of online customer reviews \citep{titov2008joint, Mei2007Topic, Lin2009JSM}.

Note that the existing works mainly focused on topic discovery and sentiment prediction using the review texts only.
While the information from the overall ratings has not been integrated to some extent.
It is seen that the rating scores provide intuitive orientations of user opinions, which can allow the latent sentiments extracted more appropriately \citep{li2015generative}.
Moreover, most collected review texts in practice are vague in the sense of low ``signal to noise ratio'' with large amounts of spam content, unhelpful opinions, as well as highly subjective and misleading information \citep{Yue2010Exploiting}.
In such situations, it is of great importance to consider ratings and review texts in a mutually complement manner for accurate quantification on review sentiments and topics.
		
The scope of this work is to predict both sentiments and topics from the joint learning of review texts and overall ratings.
Typically, the association between textual reviews and overall ratings are prevailing based on the general orientation of review sentiments.
For instance, a review stimulated by positive sentiment would present both a higher rating and a positive review text.
The sentiment polarities indicated by the overall ratings and textual reviews are closely related, while their relationship varies among different customers.
In practice, customers may have different preference and emphasis on different aspects for the same product, and they may give overall ratings based on the partial or whole product aspects discussed in review texts \citep{li2015generative}.
For example, even a full 5-star rating could be accompanied by negative review content. The dynamic relationship between the overall ratings and the review texts makes it challenging to digest the information in reviews with ratings jointly. As ratings serve as one of the most important metadata of review documents, this problem can be viewed from the perspective of the incorporation of document metadata with the content of the text \citep{roberts2016model}.
		
To address the aforementioned challenges, we propose a joint sentiment-topic model to accommodate both ratings and review texts.
We denote the proposed model as the {\it JST-RR model}.
The proposed method extends the conventional joint sentiment-topic modeling by incorporating the generative process of ratings with textual reviews in a unified framework.
Under this framework, the connection between review texts and ratings is characterized by the latent joint sentiment-topic distribution.
We have also developed a weighting mechanism between the number of of review words and the number of ratings for a more appropriate quantification on review sentiments.
The proposed JST-RR model enables an effective identification of topics and sentiments in reviews and a more accurate prediction for review data.
It is worth to pointing out that the proposed model is weakly supervised with the only supervision from a domain-independent sentiment lexicon.
Hence it can be easily adapted to review mining in various domains or applications such as political discussion in social media and detection of fake news.
		
The remainder of this paper is organized as follows. Section~\ref{literature} reviews the state-of-the-art methods on joint sentiment-topic prediction in review modeling. Section~\ref{methodology} presents the details of the proposed JST-RR model. Section~\ref{case} reports the model implementation and performance on the Amazon datasets. Section~\ref{simulation} conducts a simulation study to extensively evaluate the performance of the proposed model.
Finally, we conclude this work with some discussion in Section~\ref{conclusion}.

        \section{Literature Review}
        \label{literature}
This section mainly reviews modeling methods for online review data in sentiment-topic prediction.
In the literature, many existing works \citep{lu2009rated, brody2010unsupervised, lu2011multi} performed topic detection and sentiment classification in a two-stage process.
They first detected topics from review texts using traditional topic models such as latent Dirichlet allocation (LDA) \citep{blei2003latent} and probabilistic latent semantic indexing (PLSI) \citep{Hofmann1999Probabilistic}. Then sentiment labels are assigned to specific topics by applying sentiment classification techniques to corresponding review texts.
There are several works on detecting topics and sentiments simultaneously from user-generated content \citep{Mei2007Topic, titov2008joint, Lin2009JSM}.
For example, \cite{Mei2007Topic} proposed the topic-sentiment mixture (TSM) model for the weblog collection based on the model setting of PLSI.
However, the topic-sentiment correlation in the TSM model was not directly constructed but captured through a post-processing of model parameters.
With the focus of finding correlated sentiments and topics from texts, the joint sentiment-topic (JST) model \citep{Lin2009JSM} and the Reverse-JST model \citep{lin2012weakly} extended the LDA model by constructing an additional sentiment layer conditioning and being conditioned on the topic layer of LDA, respectively.
Many follow-up works \citep{moghaddam2011ilda, sentitopic2013, TS2015}, regarded as variants of the JST and Reverse-JST models, use the same assumption of conditional inter-dependency between topics and sentiments.
However, these works mainly focused on topic discovery and sentiment prediction from review texts only, where the information from the overall ratings has been overlooked to some extent.
		
For review sentiment prediction, existing methods often employed a supervised learning framework using sentiment labels directly indicated by overall ratings \citep{2002Thumbs,2007Biographies,Ye2009Sentiment}.
That is, the ratings were used to supervise the sentiment prediction of corresponding review texts.
However, there is still a discrepancy between the sentiment orientations indicated by review texts and ratings,
since customers may give overall ratings based on the partial or whole product aspects discussed in review texts.
Considering the complex and dynamic relationship between the overall ratings and the review texts,
it is beneficial to construct a joint model of textual reviews and numerical ratings for sentiment-topic prediction.
For instance, the models by \cite{wang2010latent,wang2011latent} assumed that the overall ratings were based on ratings of specific aspects or topics extracted from review texts. The aspect identification and rating (AIR) model by \cite{li2015generative} followed a reverse assumption that aspect ratings were produced with the prior information of overall ratings. However, these models mainly focused on the detection of aspect ratings and conditioned the joint modeling of textual reviews and overall ratings on the results of aspect ratings.

Motivated by the lack of a general model to characterize the intrinsic connection between review texts and overall ratings,
we propose a joint sentiment-topic model to accommodate both overall ratings and review texts in a unified probabilistic framework for accurate prediction of review sentiments and topics.

\section{Joint Sentiment-Topic Model of Review Texts and Ratings} \label{methodology}
In this section, we briefly describe the notation and joint sentiment-topic (JST) representation of reviews in Section \ref{JST}.
We then detail the proposed JST-RR model for integrating the overall ratings with review words in Section \ref{model}.
The procedure of model inference is constructed in Section \ref{inference}.
		
\subsection{Joint Sentiment-Topic Representation of Review Texts} \label{JST}

Consider the data consisting of a collection of product review documents $\{d_{i}, i = 1, \ldots, D\}$. For each review document $d_{i}$, suppose that it contains $N_{i}$ words denoted as $\bm w_{i} = (w_{i1}, \ldots, w_{iN_{i}})$, and it contains $M_{i}$ rating scores denoted as $\bm r_{i} = (r_{i1}, \ldots, r_{iM_{i}})$.
A review document can be composed of a single review (i.e., $M_i=1$) or a collection of reviews for learning user opinions from various granularity levels.
For example, multiple reviews of the same product or the same user can be integrated into a document for extracting the product-specific or user-specific features \citep{ling2014ratings}. Here, each word in the observed document is assumed to be from the vocabulary indexed by $\{1, \ldots, V\}$.
Without loss of generality, we assume that the rating $r_{i j} \in \{1,2,3, 4, 5\}$ with $5$ to be the highest rating and $1$ to be the lowest rating.
		
In a typical joint sentiment-topic modeling framework \citep{Lin2009JSM,lin2012weakly,sentitopic2013}, each review document $d_i$ is assumed to be represented by mixtures of sentiments and topics that are interdependent. By following the assumption in the general class of mixed membership models \citep{airoldi2010reconceptualizing,airoldi2014introduction,manrique2012estimating}, each observational unit in the document belongs to a single cluster that is represented by a specific sentiment-topic pair. Let us denote the sentiment label by $l \in \{1,\dots,S\}$ and the topic label by $z \in \{1,\dots,K\}$. The sentiment of the document $d_i$ follows a multinomial distribution $\text{Multinomial}(\boldsymbol{\pi}_i)$, where the $S$-dimension prior distribution $\boldsymbol{\pi}_i \sim \text{Dirichlet}(\boldsymbol{\gamma})$. Conditional on each sentiment label $l \in \{1,\dots,S\}$, the topic follows a multinomial distribution $\text{Multinomial}(\boldsymbol{\theta}_{i,l})$, where the $K$-dimension prior of topic distribution $\boldsymbol{\theta}_{i,l} \sim \text{Dirichlet}(\boldsymbol{\alpha}_l)$. Typically, the document-level sentiment and topic distributions indicate how likely the current document fits a specific sentiment and topic, providing a quantification on the latent sentiments and topics for unstructured reviews.
		
\subsection{Model Representation of the JST-RR Model} \label{model}

\begin{figure}[htbp]
	    \centering
	    \includegraphics[scale=0.5]{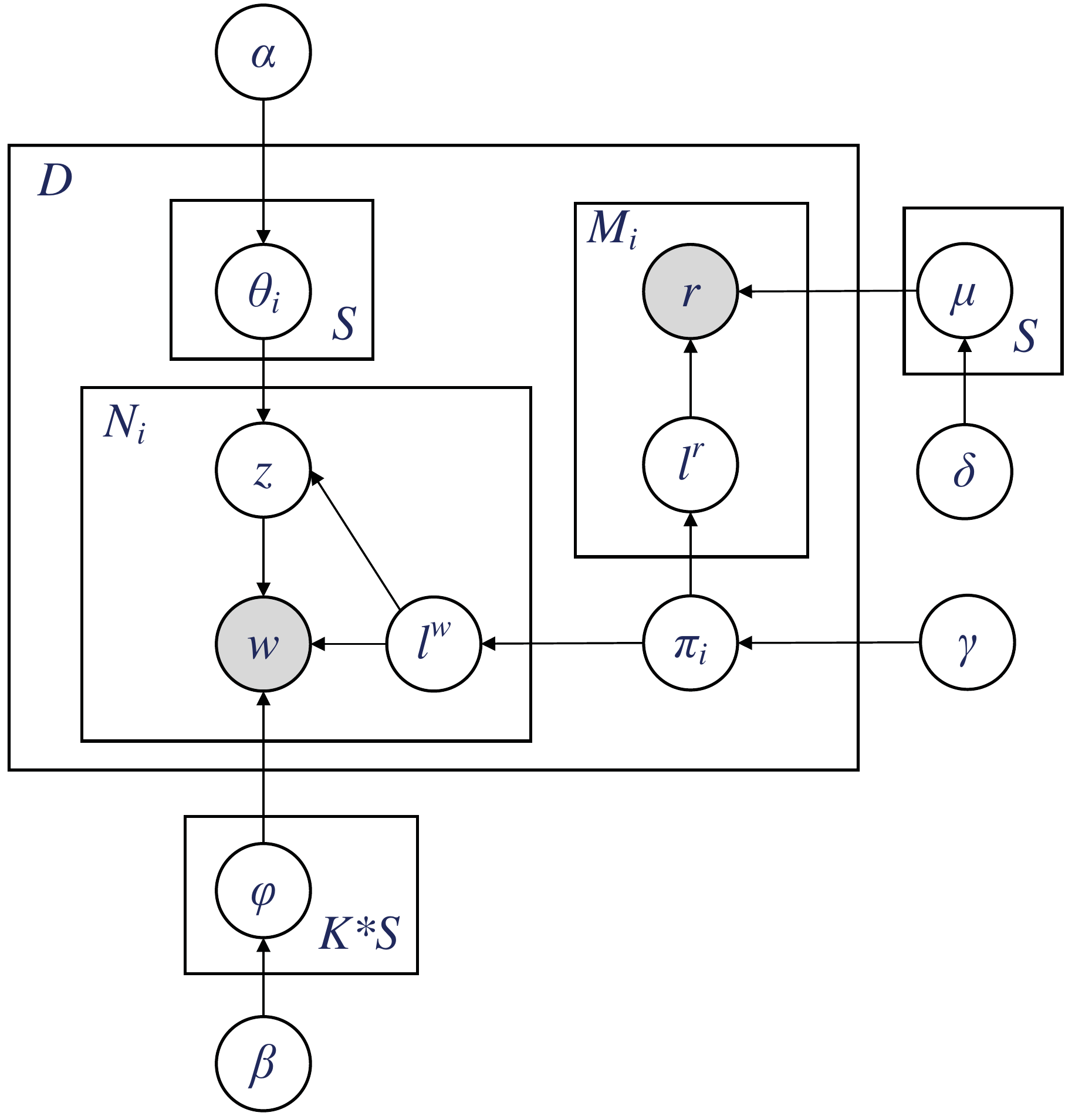}
	    \caption{Illustration of the proposed JST-RR model.} \label{fig1}
        \end{figure}

In this section, we detail the proposed JST-RR model with the consideration of both ratings and review texts.
Based on the JST representation, the proposed JST-RR model integrates the overall ratings with textual words in review documents under a unified probabilistic framework.
Figure~\ref{fig1} illustrates a graphical representation of the JST-RR model structure.
The notation used in the proposed model is summarized in Table~\ref{tab:0}.

        \begin{table}[htbp]	
        	\caption{A summary of notation.}
        	\label{tab:0}       
        	\renewcommand\arraystretch{}
        	\begin{tabular*}{1\textwidth}{p{.15\textwidth} p{.8\textwidth}}
        		\hline \hline\noalign{\smallskip}
        		Term & Definition \\
        		\noalign{\smallskip}\hline\noalign{\smallskip}
        		$d$ & Document \\
        		$w$ & Word \\
        		$r$ & Rating \\
        		$z$ & Topic label\\
        		$l^w$ & Word sentiment label \\
        		$l^r$ & Rating sentiment label \\     		
        		$D$ & Number of documents \\
        		$V$ & Vocabulary size \\        		
        		$K$ & Number of topics \\
        		$S$ & Number of sentiment classifications \\
        		\hline
        		$\boldsymbol{\pi}_i$ & Coefficient vector of the multinomial sentiment distribution for the $i$th document\\
        		$\boldsymbol{\theta}_{i,l}$ & Coefficient vector of the multinomial topic distribution under the sentiment label $l$ for the $i$th document \\        		
        		$\boldsymbol{\varphi}_{l,z}$ & Coefficient vector of the multinomial word distribution under the sentiment label $l$ and topic label $z$ \\
        		$\boldsymbol{\mu}_{l}$ & Coefficient vector of the multinomial rating distribution under the sentiment label $l$  \\
        		\hline
        		$N_i$ & Number of words in the $i$th document \\        		
        		$N_{i,l}$ & Number of words that are assigned to the sentiment label $l$ in the $i$th document \\
        		$N_{i,l,z}$ & Number of words that are assigned to the sentiment label $l$ and topic label $z$ in the $i$th document \\
        		$N_{l,z}$ & Number of words that are assigned to the sentiment label $l$ and topic label $z$ in the dataset \\
        		$N_{l,z,w}$ & Number of times that the word $w$ is assigned to the sentiment label $l$ and topic label $z$ in the dataset\\
        		$M_i$ & Number of ratings in the $i$th document \\
        		$M_{i,l}$ & Number of ratings that are assigned to the sentiment label $l$ in the $i$th document \\
        		$M_{l}$ & Number of ratings that are assigned to the sentiment label $l$ in the dataset \\
        		$M_{l,r}$ & Number of times that the rating $r$ is assigned to the sentiment label $l$ in the dataset\\        		
        		\noalign{\smallskip}\hline \hline
        	\end{tabular*}
        \end{table}

We consider that each review document $d_i$ is represented by its document-level sentiment distribution $\text{Multinomial}(\bm \pi_i)$ and topic distribution $\text{Multinomial}(\bm \theta_{i})$.
The key of the JST-RR model is to provide a unified probabilistic generative process for both observed words and ratings in the review documents.
That is, each word is assumed to be drawn from the $V$-dimension multinomial word distribution $\text{Multinomial}(\boldsymbol{\varphi}_{l^w,z})$ conditioned on the word sentiment label $l^w$ and topic label $z$, where the prior distribution $\boldsymbol{\varphi}_{l^w,z} \sim \text{Dirichlet}(\boldsymbol{\beta}_{l^w,z})$.
For the generating process of overall ratings, we consider that the overall ratings provide only a general orientation of sentiments.
Each rating is then assumed to be drawn from the $5$-dimension multinomial rating distribution $\text{Multinomial}(\boldsymbol{\mu}_{l^r})$ only conditioned on its rating sentiment label $l^r$, where the prior distribution $\boldsymbol{\mu}_{l^r}\sim\text{Dirichlet}(\boldsymbol{\delta}_{l^r})$.

        A formal generative process of the review document collection $\{d_{i}, i = 1, \ldots, D\}$ is presented in Procedure \ref{alg1}.
        In this framework, words and ratings are jointly generated and used as observations for the estimation of reviews.
        The hyperparameters $\boldsymbol{\beta}$, $\boldsymbol{\delta}$, $\boldsymbol{\gamma}$, and $\boldsymbol{\alpha}$ indicate the prior information before the actual words and ratings, i.e., the actual data, are observed.
        The settings of hyperparameters are detailed in Section~\ref{case-1} based on a real-world case.

        \begin{algorithm}[ht]
        \caption{Generative procedure of words and ratings of the JST-RR model}
        \label{alg1}
        \begin{itemize}[leftmargin=*]
        \item For the entire document collection, first characterize the ``topic'' and the ``sentiment'' by the word probability distribution and the rating probability distribution:
        \begin{itemize}	
        	\item For each combination of word sentiment label $l^w\in\{1,\dots,S\}$ and topic label $z\in\{1,\dots,K\}$:
        	\begin{itemize}
        		\item Draw sample from word probability distribution $\boldsymbol{\varphi}_{l^w,z}\sim\text{Dirichlet}(\boldsymbol{\beta}_{l^w,z})$.
        	\end{itemize}
        	\item For each rating sentiment label $l^r\in\{1,\dots,S\}$:
        	\begin{itemize}
        		\item Draw sample from rating probability distribution $\boldsymbol{\mu}_{l^r}\sim\text{Dirichlet}(\boldsymbol{\delta}_{l^r})$.
        	\end{itemize}
        	\end{itemize}
      	
        \item For each document $d_{i}, i = 1, \ldots, D$:
        	\begin{itemize}
        		\item Draw sample from sentiment probability distribution $\boldsymbol{\pi}_i\sim\text{Dirichlet}(\boldsymbol{\gamma})$.
        		\item Draw sample from topic probability distribution $\boldsymbol{\theta}_{i,l}\sim\text{Dirichlet}(\boldsymbol{\alpha}_l)$ for each sentiment label $l\in\{1,\dots,S\}$.
        		\item For each word $w_{ij},j=1,\dots,N_i$ in document $d_i$:
        		\begin{itemize}
        			\item Draw the sentiment assignment $l^w_{ij}\sim\text{Multinomial}(\boldsymbol{\pi}_i)$.
        			\item Draw the topic assignment $z_{ij}\sim\text{Multinomial}(\boldsymbol{\theta}_{i,l^w_{ij}})$ conditioned on $l^w_{ij}$.
        			\item Draw a specific word $w_{ij}\sim\text{Multinomial}(\boldsymbol{\varphi}_{l^w_{ij},z_{ij}})$ conditioned on $l^w_{ij}$ and $z_{ij}$.
        		\end{itemize}
        		\item For each rating $r_{ij},j=1,\dots,M_i$ in document $d_i$:
        		\begin{itemize}
        			\item Draw the sentiment assignment $l^r_{ij}\sim\text{Multinomial}(\boldsymbol{\pi}_i)$.
        			\item Draw a specific rating score $r_{ij}\sim\text{Multinomial}(\boldsymbol{\mu}_{l^r_{ij}})$ conditioned on $l^r_{ij}$.
        		\end{itemize}
        	\end{itemize}
        	\end{itemize}
        \end{algorithm}

The proposed JST-RR model not only provides a probabilistic and unified framework, but also provides a meaningful manner on how ratings and review texts work in realistic settings.
For example, a reviewer on Amazon has an overall sentiment regarding the purchased product, which informs the reviewer's sentiment on the various aspects of the product which are typically represented as ``topics'' in the model.
It is likely that the reviewer has a negative sentiment about one topic while having a positive sentiment on other topics,
and this can be reflected by the word sentiment and overall sentiment from the proposed model.
		
\subsection{Model Inference of the JST-RR Model} \label{inference}
For the inference of the proposed JST-RR model, there are four sets of latent distribution parameters:  the document-level sentiment distribution parameter $\bm \pi$, the sentiment-specific topic distribution parameter $\bm \theta$, the joint sentiment/topic-word distribution parameter $\bm \varphi$, and the sentiment-rating distribution parameter $\bm \mu$.
Given these latent distributions,
we can explicitly express the joint probability of the observed words, ratings, and their sentiment/topic labels in the document collection $\{d_i,i=1,\dots,D\}$ as
		\begin{align}
		\begin{split}
		    &P(\bm w, \bm r,\bm l^w, \bm l^r, \bm z | \bm \pi,\bm \theta,\bm \varphi,\bm \mu) = \prod_{i=1}^{D} \prod_{j=1}^{N_i}P(l^w_{ij},z_{ij},w_{ij}|\bm\pi_i,\bm\theta_{i,l^w_{ij}},\bm \varphi_{l^w_{ij},z_{ij}}) \prod_{j=1}^{M_i}P(l^r_{ij},r_{ij}|\bm\pi_i,\bm\mu_{l^r_{ij}})\\
		    &= \prod_{i=1}^{D} \prod_{j=1}^{N_i}P(l^w_{ij}|\bm\pi_i)P(z_{ij}|\bm\theta_{i,l^w_{ij}})P(w_{ij}|\bm \varphi_{l^w_{ij},z_{ij}}) \prod_{j=1}^{M_i}P(l^r_{ij}|\bm \pi_i)P(r_{ij}|\bm\mu_{l^r_{ij}}),
		\end{split}
		\end{align}
where the words and ratings are conditionally independent given the document-level sentiments and topics.
It is seen that the observed words are dependent on their latent sentiment and topic assignments, while the ratings are only dependent on their latent sentiment assignments.
		
Note that there have been several methods in the literature developed for the inference of probabilistic topic models, including Gibbs sampling \citep{griffiths2004finding}, variational Bayesian inference \citep{blei2003latent}, and maximum a posteriori (MAP) estimation \citep{chien2008MAP}.
In this work, we adopt the Gibbs sampling for the model inference because of its promising convergence to the underlying distribution.
It is also noted that some advanced algorithms \citep{hoffman2013stochastic,srivastava2017autoencoding} could be adapted to our problem for handling large and complex data.
The state transition of the Markov chain formed by the Gibbs sampler is determined by the sampling of the latent variables (i.e., the topic label $z$ and the sentiment label $l$) given the current values of all other variables and the observed data.
The conditional probability of sampling sentiment label $l^w_{ij}$ and topic label $z_{ij}$ for the observed word $w_{ij}=w$ in the document $d_i$ can be written as
		\begin{align} \label{eq:JSTRR-1}
		\begin{split}
		&P(l^w_{ij}=l,z_{ij}=z|\bm{w},\bm{l}^w_{-ij},\bm{l}^r,\bm{z}_{-ij}) \propto P(l^w_{ij}=l,z_{ij}=z,w_{ij}=w|\bm{w}_{-ij},\bm{l}^w_{-ij},\bm{l}^r,\bm{z}_{-ij})\\
		&=P(l^w_{ij}=l|\bm{l}^w_{-ij},\bm{l}^r)\times P(z_{ij}=z|l^w_{ij}=l,\bm{l}^w_{-ij},\bm{z}_{-ij})\times P(w_{ij}=w|l^w_{ij}=l,z_{ij}=z,\bm{l}^w_{-ij},\bm{z}_{-ij},\bm{w}_{-ij})\\
		&=\int_{\bm{\pi}_i} P(l^w_{ij}=l|\bm{\pi}_i)P(\bm{\pi}_i|\bm{l}^w_{-ij},\bm{l}^r)\,\mathrm{d}\bm{\pi}_i \times \int_{\bm{\theta}_{i,l}} P(z_{ij}=z|\bm{\theta}_{i,l})P(\bm{\theta}_{i,l}|\bm{l}^w_{-ij},\bm{z}_{-ij})\,\mathrm{d}\bm{\theta}_{i,l} \times \\
		&\int_{\bm{\varphi}_{l,z}} P(w_{ij}=w|\bm{\varphi}_{l,z})P(\bm{\varphi}_{l,z}|\bm{l}^w_{-ij},\bm{z}_{-ij},\bm{w}_{-ij})\,\mathrm{d}\bm{\varphi}_{l,z}.
		\end{split}		
		\end{align}
The superscript or subscript $-{ij}$ hereafter denotes the data quantity excluding the $j$th position in the document $d_i$.
By integrating out $\boldsymbol{\pi}_i$ (see detailed derivation in Appendix), the first term in Eq.~(\ref{eq:JSTRR-1}) can be derived as
	    \begin{equation}
	    \label{eq:JSTRR-2}
	    P(l^w_{ij}=l|\bm{l}^w_{-ij},\bm{l}^r)=\frac{N_{i,l}^{-ij}+M_{i,l}+\gamma_{l}}{N_i^{-ij}+M_i+\sum_{l'}\gamma_{l'}}.
	    \end{equation}
It represents the probability of sampling $l^w_{ij}=l$ given all other sentiment assignments $\bm{l}^w_{-ij}$ of words and $\bm{l}^r$ of ratings in the same review document $d_i$.
Here $N_i$ and $M_i$ are the total number of words and ratings in the document $d_i$, $N_{i,l}$ and $M_{i,l}$ are the number of words and ratings associated with sentiment $l$ in the document $d_i$.
The hyperparameter $\gamma_l$ can be interpreted as the prior observation counts of the sentiment $l$ assigned with $d_i$.
	
From Eq.~(\ref{eq:JSTRR-2}) and its derivation in the Appendix, one can see that the number of ratings and the number of words are treated with the equal weight for the estimation of sentiments.
However, in a typical user review,
the number of words is often much larger than the number of ratings, even when multiple ratings are allowed in a particular application.
Moreover, the number of ratings and the number of words associated with a specific sentiment (i.e., $N_{i,l}$ and $M_{i,l}$) in Eq.~(\ref{eq:JSTRR-2}) are not generated from actual i.i.d. observations but from the results of a Gibbs sampler.
The value of $N_{i,l}$ and $M_{i,l}$ reflects different confidence levels, i.e., the  number of words used to express a particular opinion in a review is large relative to the sentiments expressed as ratings.
To address these challenges, we consider to incorporate a weighting mechanism between the number of ratings and the number of words in sentiment estimation.
From the perspective of a weighted likelihood for the sentiment assignments of words and ratings (see detailed derivation in Appendix),
Eq.~(\ref{eq:JSTRR-2}) can be re-expressed in a more general form:
		\begin{equation}
		\label{eq:JSTRR-3}
		P(l^w_{ij}=l|\bm{l}^w_{-ij},\bm{l}^r)=\frac{N_{i,l}^{-ij}+\sigma M_{i,l}+\gamma_{l}}{N_i^{-ij}+\sigma M_i+\sum_{l'}\gamma_{l'}},
		\end{equation}
where $\sigma$ is a weighting parameter to indicate the weight of a rating relative to a word in the estimation of review sentiments.
When $\sigma=0$, the document-level sentiment prediction depends only on the review words, which is simplified as the JST model in \cite{Lin2009JSM}.

Similarly, the second term in Eq.~(\ref{eq:JSTRR-1}) can be estimated by integrating out $\boldsymbol{\theta}_{i,l}$, which gives
	    \begin{equation}
	    \label{eq:JSTRR-4}
	    P(z_{ij}=z|l^w_{ij}=l,\bm{l}^w_{-ij},\bm{z}_{-ij})=\frac{N^{-ij}_{i,l,z}+\alpha_{l,z}}{N^{-ij}_{i,l}+\sum_{z'}\alpha_{l,z'}},
	    \end{equation}
where $N_{i,l,z}$ is the number of words associated with the sentiment $l$ and topic $z$ in the document $d_i$, and the hyperparameter $\alpha_{l,z}$ can be interpreted as the prior observation counts of words assigned with the sentiment $l$ and topic $z$ in $d_i$.
For the third term in Eq.~(\ref{eq:JSTRR-1}), we can obtain its posterior prediction
by integrating out $\boldsymbol{\varphi}_{l,z}$ in the same manner as
	    \begin{equation}
	    \label{eq:JSTRR-5}
	    P(w_{ij}=w|l^w_{ij}=l,z_{ij}=z,\bm{l}^w_{-ij},\bm{z}_{-ij},\bm{w}_{-ij})=\frac{N_{l,z,w}^{-ij}+\beta_{l,z,w}}{N_{l,z}^{-ij}+\sum_{w'}\beta_{l,z,w'}},
	    \end{equation}
where $N_{l,z}$ is the number of words assigned with the sentiment label $l$ and topic label $z$ in the entire dataset, $N_{l,z,w}$ is the number of times that the word $w$ is associated with the sentiment label $l$ and topic label $z$ in the dataset, and the hyperparameter $\beta_{l,z,w}$ can be interpreted as the prior counts of word $w$ associated with sentiment label $l$ and topic label $z$ in the dataset.
	
By combining the results in Eq.~(\ref{eq:JSTRR-3}), Eq.~(\ref{eq:JSTRR-4}), and Eq.~(\ref{eq:JSTRR-5}),
the expression for the full conditional probability in Eq.~(\ref{eq:JSTRR-1}) can be written as
	    \begin{equation}
	    \label{eq:JSTRR-6}
	    P(l^w_{ij}=l,z_{ij}=z|\bm{w},\bm{l}^w_{-ij},\bm{l}^r,\bm{z}_{-ij})\propto\frac{N_{i,l}^{-ij}+\sigma M_{i,l}+\gamma_{l}}{N_i^{-ij}+\sigma M_i+\sum_{l'}\gamma_{l'}}\cdot\frac{N^{-ij}_{i,l,z}+\alpha_{l,z}}{N^{-ij}_{i,l}+\sum_{z'}\alpha_{l,z'}}\cdot\frac{N_{l,z,w}^{-ij}+\beta_{l,z,w}}{N_{l,z}^{-ij}+\sum_{w'}\beta_{l,z,w'}}.
	    \end{equation}
In a similar manner,
we can specify the conditional probability of sampling the sentiment label $l^r_{ij}$ for the observed rating $r_{ij}=r$ in the document $d_i$ as (see detailed derivation in Appendix)
		\begin{equation}
		\label{eq:JSTRR-7}
		\begin{split}
		&P(l^r_{ij}=l|\bm{r},\bm{l}^r_{-ij},\bm{l}^w)\propto P(l^r_{ij}=l,r_{ij}=r|\bm{r}_{-ij},\bm{l}^r_{-ij},\bm{l}^w)\\
		&=P(l^r_{ij}=l|\bm{l}^r_{-ij},\bm{l}^w)\times P(r_{ij}=r|l^r_{ij}=l,\bm{l}^r_{-ij},\bm{r}_{-ij})\\
		&=\int_{\bm{\pi}_i} P(l^r_{ij}=l|\bm{\pi}_i)P(\bm{\pi}_i|\bm{l}^r_{-ij},\bm{l}^w)\,\mathrm{d}\bm{\pi}_i \times \int_{\bm{\mu}_{l}} P(r_{ij}=r|\bm{\mu}_{l})P(\bm{\mu}_{l}|\bm{l}^r_{-ij},\bm{r}_{-ij})\,\mathrm{d}\bm{\mu}_{l}\\
		&=\frac{N_{i,l}+\sigma M_{i,l}^{-ij}+\gamma_{l}}{N_{i}+\sigma M_{i}^{-ij}+\sum_{l'}\gamma_{l'}}\times\frac{M_{l,r}^{-ij}+\delta_{l,r}}{M_{l}^{-ij}+\sum_{r'}\delta_{l,r'}},
		\end{split}
		\end{equation}
where $M_{l}$ is the number of ratings associated with the sentiment $l$ in the dataset, $M_{l,r}$ is the number of times that the rating $r$ is associated with sentiment label $l$ in the dataset, and the hyperparameter $\delta_{l,r}$ can be interpreted as the prior counts of rating $r$ associated with sentiment label $l$ in the dataset.
		
A sample obtained from the Markov chain in its stable state is used to obtain the posterior estimations of the parameters $\boldsymbol{\pi}$, $\boldsymbol{\theta}$, $\boldsymbol{\varphi}$, and $\boldsymbol{\mu}$ as follows:
		\begin{equation}
		\label{eq:JSTRR-8}
		\begin{split}
		\hat{\pi}_{i,l}=\frac{N_{i,l}+\sigma M_{i,l}+\gamma_{l}}{N_i+\sigma M_i+\sum_{l'}\gamma_{l'}},\quad \hat{\theta}_{i,l,z}=\frac{N_{i,l,z}+\alpha_{l,z}}{N_{i,l}+\sum_{z'}\alpha_{l,z'}},\\
		\hat{\varphi}_{l,z,w}=\frac{N_{l,z,w}+\beta_{l,z,w}}{N_{l,z}+\sum_{w'}\beta_{l,z,w'}},\quad \hat{\mu}_{l,r}=\frac{M_{l,r}+\delta_{l,r}}{M_{l}+\sum_{r'}\delta_{l,r'}}.
		\end{split}
		\end{equation}
For each document $d_i$, its document-level sentiment distribution parameter $\boldsymbol{\pi}_i$ is approximated based on both $N_i$ words and $M_i$ ratings with a weighting parameter $\sigma$, while the topic distribution parameter $\boldsymbol{\theta}_i$ is estimated by only words in the document since ratings are not assigned with topic labels.
The Gibbs sampling procedure of making inference of the proposed JST-RR model is summarized in Procedure \ref{alg2}.
		
		\begin{algorithm} [ht]
        \caption{Gibbs sampling procedure for inference of the JST-RR model}
        \label{alg2}
        \KwIn{Document collection $\{d_i,\,i=1,\dots,D\}$, hyperparameters $\boldsymbol{\beta}$, $\boldsymbol{\delta}$, $\boldsymbol{\gamma}$, $\boldsymbol{\alpha}$, and weight parameter $\sigma$.}
        \KwOut{Word distribution parameter $\boldsymbol{\varphi}$, rating distribution parameter $\boldsymbol{\mu}$, document-level sentiment distribution parameter $\boldsymbol{\pi}$ and topic distribution parameter $\boldsymbol{\theta}$.}
        \BlankLine
        Assign initial topic/sentiment labels to all words/ratings at random\;
        \For{each Gibbs sampling iteration}
        {
            \For{each document $d_i,i=1,\dots,D$}
            {
                \For{each word $w_{ij},j=1,\dots,N_i$ in the document $d_i$}
                {
                    Exclude $w_{ij}$ associated with its sentiment label $l^w_{ij}$ and topic label $z_{ij}$ from count variables $N_{i}$, $N_{i,l}$, $N_{i,l,z}$, $N_{l,z}$, $N_{l,z,w}$\;
				    Sample a new sentiment-topic combination for $w_{ij}$ based on Eq.~(\ref{eq:JSTRR-6})\;
				    Update count variables $N_{i}$, $N_{i,l}$, $N_{i,l,z}$, $N_{l,z}$, $N_{l,z,w}$ by incorporating the new sentiment/topic label of $w_{ij}$\;
                }
                \For{each rating $r_{ij},j=1,\dots,M_i$ in the document $d_i$}
                {
                    Exclude $r_{ij}$ associated with its sentiment label $l^r_{ij}$ from count variables $M_{l}$, $M_{l,r}$, $M_{i}$, $M_{i,l}$\;
				    Sample a new sentiment assignment for $r_{ij}$ based on Eq.~(\ref{eq:JSTRR-7})\;
				    Update count variables $M_{l}$, $M_{l,r}$, $M_{i}$, $M_{i,l}$ by incorporating the new sentiment label of $r_{ij}$\;
                }
            }
        }
        Estimate $\boldsymbol{\varphi}$, $\boldsymbol{\mu}$, $\boldsymbol{\pi}$, and $\boldsymbol{\theta}$ based on Eq.~(\ref{eq:JSTRR-8})\;
        \end{algorithm}

\section{Case Study of Amazon Datasets} \label{case}

In this section, we evaluate the performance of the proposed model using three real datasets.
The real data are obtained from the publicly available Amazon datasets \citep{Mcauley2015Image}.
Specifically, the three datasets are the online reviews of HP laptops, the online reviews of Lenovo laptops, and the online reviews of Dell laptops, which are denoted as \textit{HP}, \textit{Lenovo}, and \textit{Dell}, respectively.
For each single review, there is an overall rating that ranges from $1$ star to $5$ stars.
	
By defining the review documents at various granularity levels (i.e. from a single review, to a collection of reviews from the same product or the same user),
the proposed JST-RR model can be applied for modeling customer opinions on different levels of interest.
In this section, we mainly focus on examining the performance of the proposed method for the individual review documents.
That is, each document here is based on a single review including a review text and an overall rating.
	
\subsection{Data Preparation and Experiment Settings} \label{case-1}

For each dataset, we perform data pre-processing in the following steps. First, we convert words into lower cases and remove the punctuation, stop words (e.g., "a", "and", "be"), and infrequent words. Second, we stem each word to its root with Porter Stemmer (http://tartarus.org/martin/PorterStemmer/). Third, we perform \emph{Negation} by adding a prefix ``not\_'' to the word in negative dependency. For example, in the sentence ``I do not like this product'', ``not\_like'' is recognized as a whole to express negative sentiment. Finally, to obtain unbiased training results on sentiment prediction, we balance the number of positive and negative review documents in the dataset. After data pre-processing, the summary statistics of three experimental datasets are listed in Table~\ref{tab:1}. For each dataset, we partition it into a $90\%$ training set for model training and a $10\%$ test set for model evaluation.
        \begin{table}[htbp]
	    	\centering
	    	\caption{A description of three Amazon datasets.}
	    	\label{tab:1}
	    	\renewcommand\arraystretch{1.5}
	    	\begin{tabular}{rcc}
	    		\hline \hline
	    		Dataset & Number of Reviews &  Average number of words (review length)\\
	    		\hline
	    		\textit{HP} & 11,655 & 71.56 \\
	    		\textit{Lenovo} & 4,976 & 71.71 \\
	    		\textit{Dell} & 8,438 & 51.09 \\
	    		\hline \hline
	    	\end{tabular}
	    \end{table}

In the implementation of the proposed method, we set the number of sentiment polarities $S=2$ (i.e., positive and negative) and a varying number of topics $K\in\{2,5,7,10,12,15,20\}$. For the setting of hyperparameters, we simply use a symmetric setting for $\boldsymbol{\gamma}$ and $\boldsymbol{\alpha}$: $\gamma_l=3.0/S,\,l\in\{1,\dots,S\}$; $\alpha_{l,z}=3.0/(S\times K),\,l\in\{1,\dots,S\},\,z\in\{1,\dots,K\}$. Based on the prior knowledge that a positive polarity is linked to a higher rating score and vice versa,  we set $\delta_{l,r}=10.0\times r,\,r\in\{1,2,3,4,5\}$ for the positive sentiment $l$, and set $\delta_{l,r}=10.0\times (6-r),\,r\in\{1,2,3,4,5\}$ for the negative sentiment $l$.

For the setting of hyperparameter $\boldsymbol{\beta}$, we use an asymmetric prior setting of $\boldsymbol{\beta}$ for the sentiment-word distribution. Note that many words are commonly treated as positive (e.g., ``excellent'') or negative (e.g., ``terrible'') regardless of the topics involved. Specifically, we select 1,048 positive words and 2,149 negative words from the sentiment lexicon MPQA (http://www.cs.pitt.edu/mpqa/) whose polarity orientations are domain independent. For the positive sentiment $l$, we set elements in $\boldsymbol{\beta}_l$ to be $0$ for the words in negative list, $0.01$ for other words. Similarly, for the negative sentiment $l$, we set elements of $\boldsymbol{\beta}_l$ to be $0$ for the words in positive list, $0.01$ for other words. Such a setting of $\boldsymbol{\beta}$ enables that the words in sentiment lexicons can only be drawn from the word distributions conditioned on their corresponding sentiment labels.

\subsection{Quantitative Performance Analysis} \label{case-2}

The proposed JST-RR model is compared with four alternative methods: JST, RJST, AIR-JST, and AIR-RJST.
The JST model in \cite{Lin2009JSM} can be treated as a baseline method for modeling topics and sentiments jointly via review texts alone.
The RJST (or Reverse-JST) method in \cite{lin2012weakly} is a variant of JST model where the topic and the sentiment layers are inverted.
The last two methods in comparison are denoted as AIR-JST and AIR-RJST based on the related AIR method in \cite{li2015generative}.
The AIR method models observed textual reviews and overall ratings in a generative way by sampling latent sentiments of review texts with the overall ratings as prior parameters.
For example, the review sentiment probability $\boldsymbol{\pi}$ is generated in accordance with its normalized rating $r$ by:
        \begin{align*}
            \boldsymbol{\pi}\sim\text{Beta}(\lambda r,\lambda(1-r)).
        \end{align*}
The AIR model is adapted to our experimental settings in this case with two variants: AIR-JST and AIR-RJST, where the sentiment and the topic layers in the two models are inverted.

To quantitatively evaluate the performance of the proposed method, we consider the perplexity based performance measure on the test set.
The perplexity is a conventional metric for evaluating the performance of probabilistic topic models.
Specifically, for a test set of documents $\{d_i,i=1,\dots,D\}$, the perplexity of observed words $\{\bm w_i,\,i=1,\dots,D\}$ in the test set is defined as:
	    \begin{equation}
	    \label{eq:case-1}
	    perplexity(\{\bm w_i,\,i=1,\dots,D\}|\boldsymbol{ \hat{\varphi}})= \exp \left \{-\frac{\sum_{i=1}^{D}\log P(\bm{w}_i|\boldsymbol{\hat{\varphi}})}{\sum_{i=1}^{D}N_i} \right \},
	    \end{equation}
where the trained model is described by the word distribution parameter $\boldsymbol{ \hat{\varphi}}$ that is estimated from the training set.
We employed the importance sampling methods in \cite{wallach2009evaluation} to approximate the probability of the observed words $P(\bm{w}_i|\boldsymbol{\hat{\varphi}})$ in Eq.~(\ref{eq:case-1}). 
Since the perplexity values monotonically decrease with the log-likelihood of the test data,
a lower perplexity indicates better prediction performance of the proposed model \citep{blei2003latent}.
It is noted that the upper bound of the perplexity in Eq.~(\ref{eq:case-1}) with the worst case of a random prediction is given by
	    \begin{equation*}
	        perplexity(\{\bm{w}_i,i=1,\dots,D\}) = \exp\left\{-\sum_{w\in V}P(w)\log P(w)\right\},
	    \end{equation*}
which is determined by the information entropy of words in the test data.
Similarly, the perplexity of observed ratings $\{\bm r_i,\,i=1,\dots,D\}$ in the test set can be defined accordingly.
As the other four models in comparison only consider the generative process of the observed words,
we conduct evaluation mainly based on the word perplexity values.


        \begin{figure}[htbp]
	    \centering
	    \includegraphics[scale=0.5]{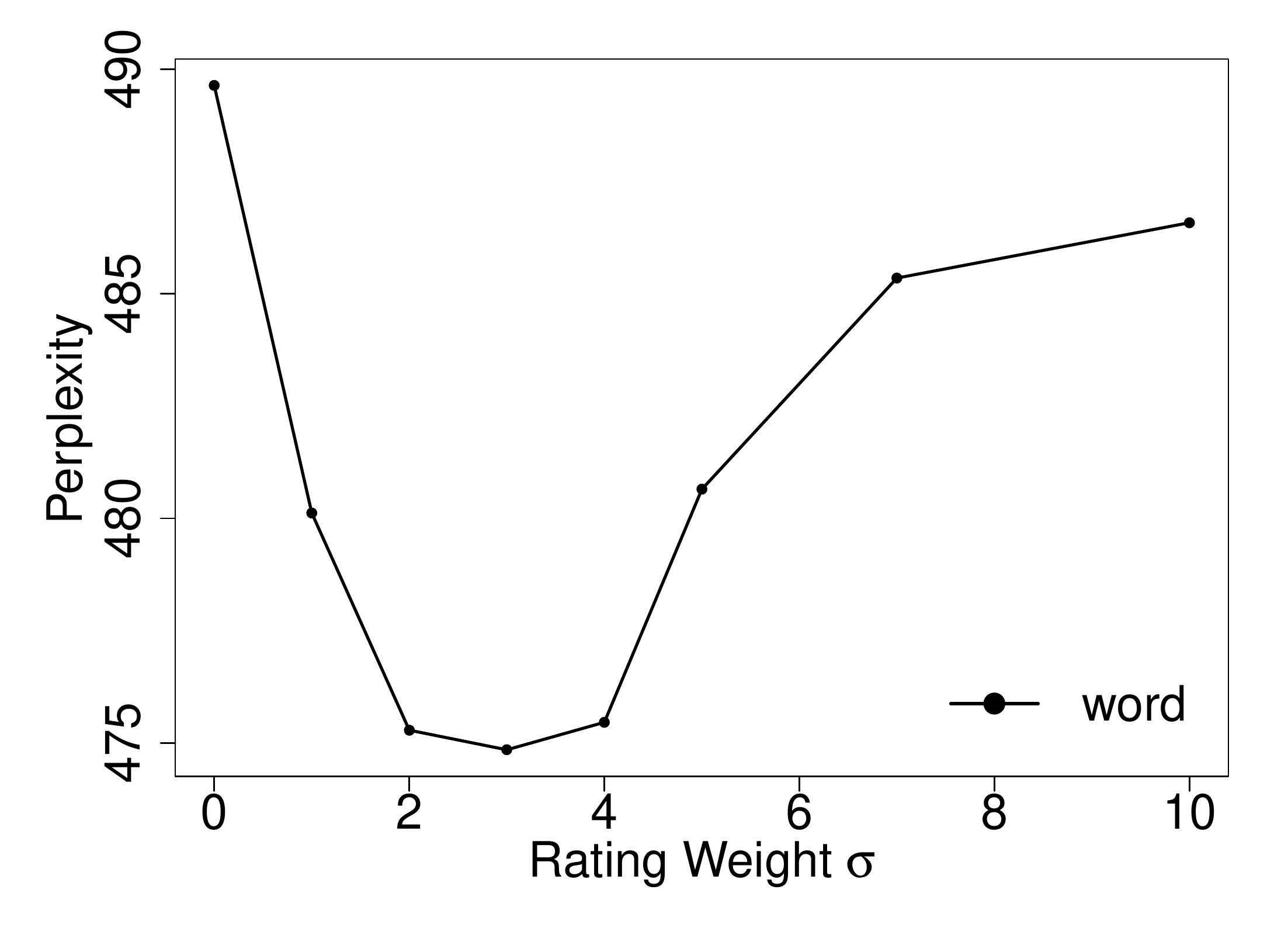}
	    \caption{The word perplexity in 10-fold cross validation for the \textit{HP} dataset with topic number $K=5$.}
	    \label{fig2}
        \end{figure}
For the selection of the tuning parameter $\sigma$ in the JST-RR model and the prior weight $\lambda$ in the two AIR models,
we adopt the 10-fold cross validation on the training set, such that the selected parameters give the average best goodness of fit (indicated by the lowest perplexity values in this study).
For example, Figure~\ref{fig2} shows the perplexity values of observed words versus the weight parameter $\sigma$ by implementing the JST-RR model in 10-fold cross validation for the \textit{HP} dataset with topic number $K=5$. Similar trends of perplexity are also observed in the other cases, and thus omitted here. Generally, a lower perplexity value indicates better model performance in explaining the observed data.
When $\sigma=0$, the proposed JST-RR model converges to the baseline JST model that only focuses on review words.
Based on the results in Figure~\ref{fig2}, the model explanation of observed words would benefit from the incorporation of ratings with a proper setting of rating weight $\sigma$.

\begin{figure}[htbp]
	    	\centering
	    	\subfigure[Perplexity on \textit{Lenovo}]{
	    		\label{fig3-a}
	    		\begin{minipage}{0.47\linewidth}
	    			\centering
	    			\includegraphics[scale=0.75]{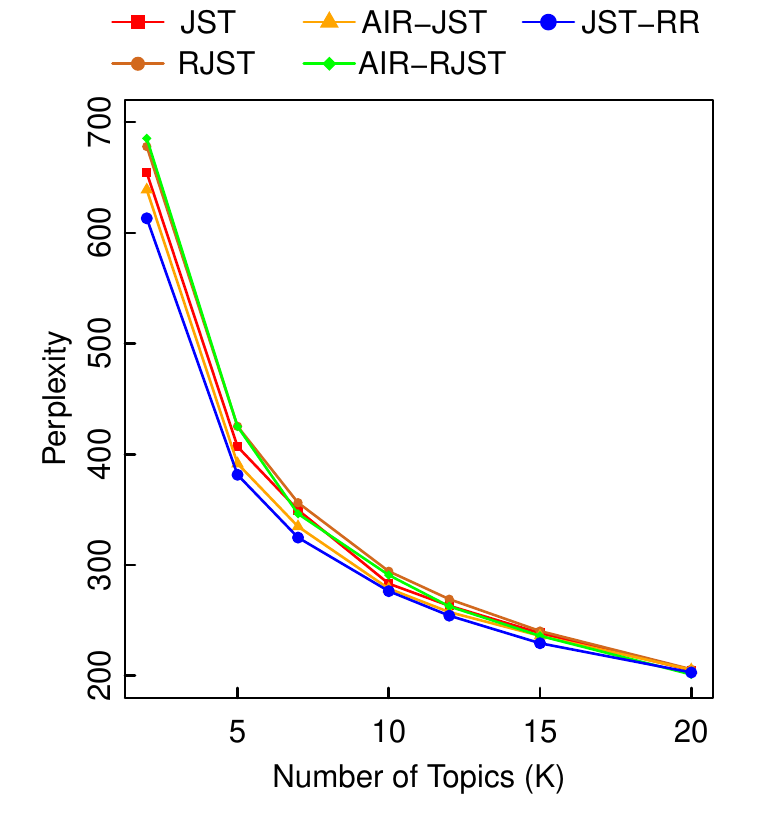}
	    		\end{minipage}
	    	}     	
	    	\subfigure[Percentage of perplexity against RJST on \textit{Lenovo}]{
	    		\label{fig3-b}
	    		\begin{minipage}{0.47\linewidth}
	    			\centering
	    			\includegraphics[scale=0.75]{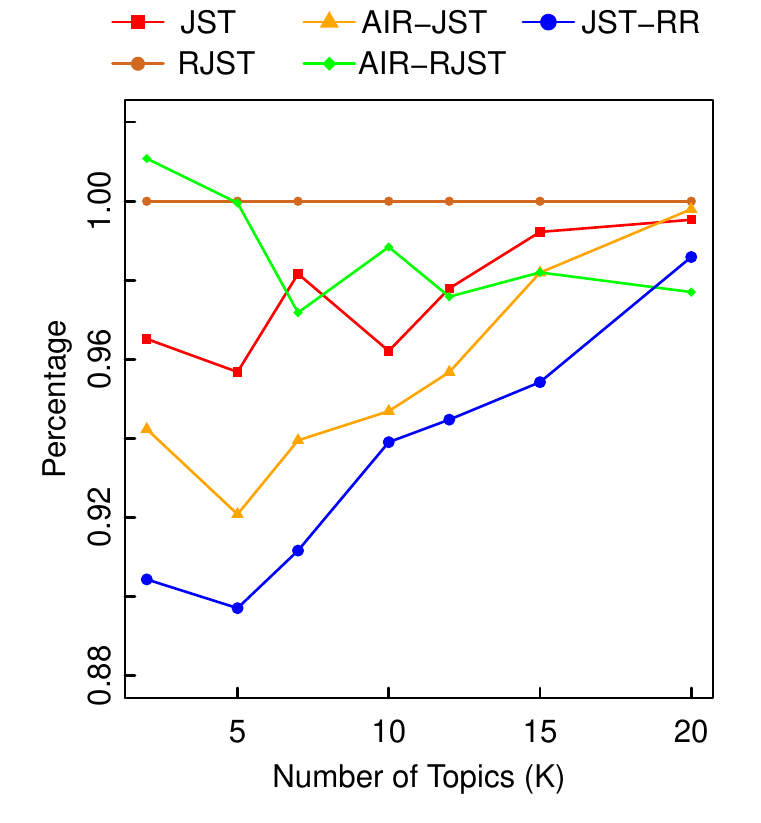}
	    		\end{minipage}
	    	}
	    	\subfigure[Perplexity on \textit{Dell}]{
	    		\label{fig3-c}
	    		\begin{minipage}{0.47\linewidth}
	    			\centering
	    			\includegraphics[scale=0.75]{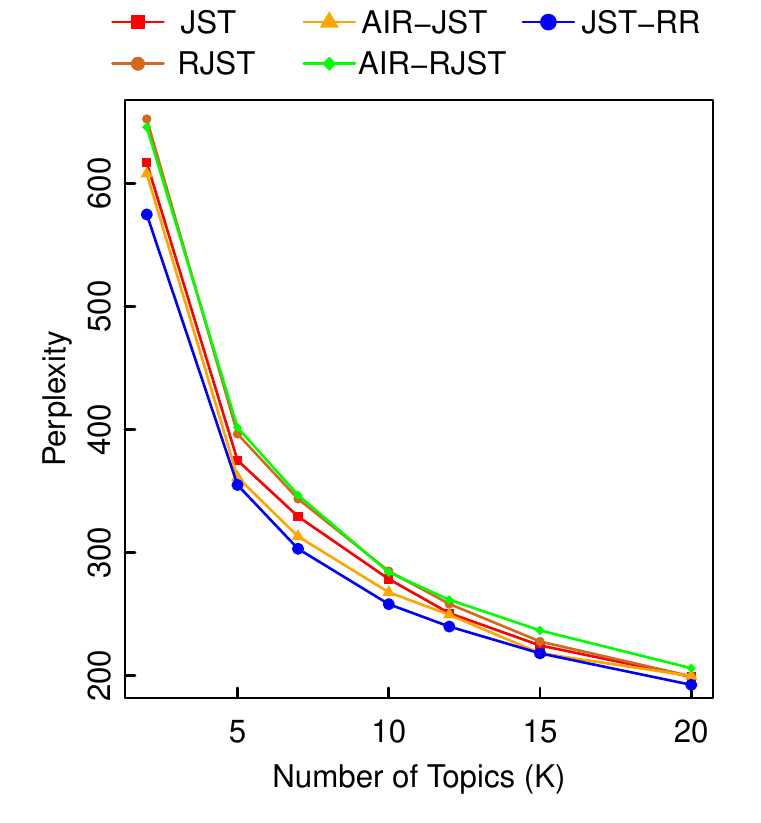}
	    		\end{minipage}
	    	}
	    	\subfigure[Percentage of perplexity against RJST on \textit{Dell}]{
	    		\label{fig3-d}
	    		\begin{minipage}{0.47\linewidth}
	    			\centering
	    			\includegraphics[scale=0.75]{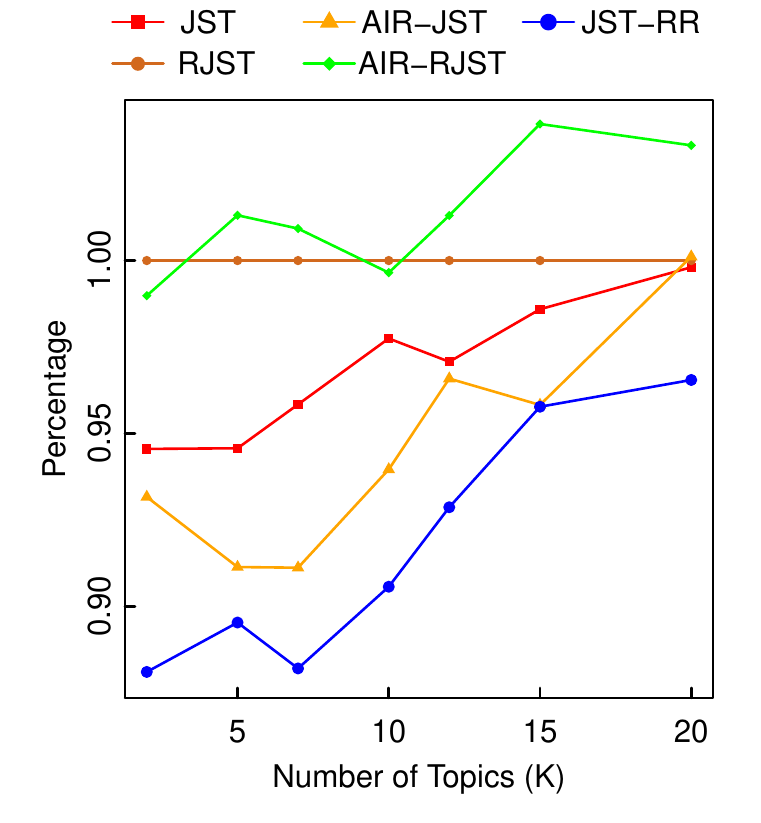}
	    		\end{minipage}
	    	}
	    	\subfigure[Perplexity on \textit{HP}]{
	    		\label{fig3-e}
	    		\begin{minipage}{0.47\linewidth}
	    			\centering
	    			\includegraphics[scale=0.75]{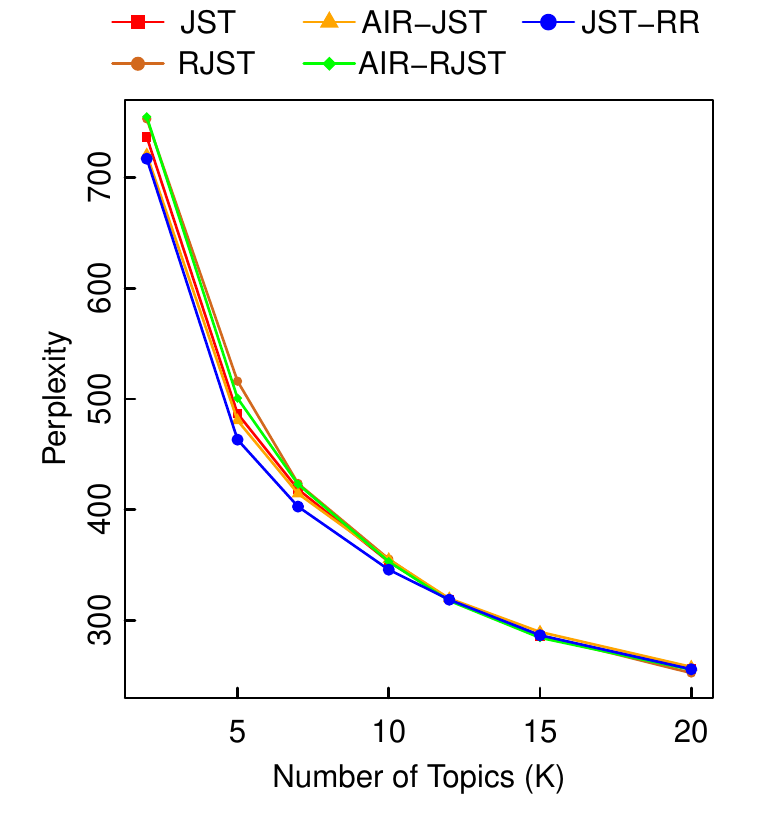}
	    		\end{minipage}
	    	}
	    	\subfigure[Percentage of perplexity against RJST on \textit{HP}]{
	    		\label{fig3-f}
	    		\begin{minipage}{0.47\linewidth}
	    			\centering
	    			\includegraphics[scale=0.75]{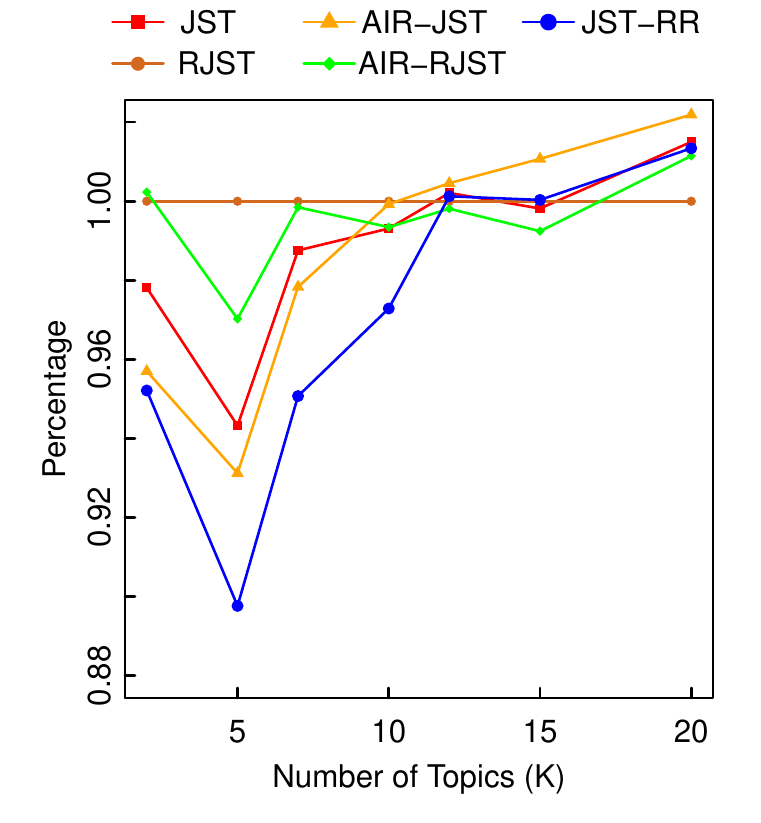}
	    		\end{minipage}
	    	}
	    	\caption{The results of word perplexity (smaller value indicating better performance) for five methods in comparison on three Amazon datasets: \textit{Lenovo}, \textit{Dell}, \textit{HP}.}
	    	\label{fig3}
	    \end{figure}

Figure~\ref{fig3} shows the word perplexity results of the five models as well as their percentages against the baseline of RJST model with varying topic numbers in comparison.
It is seen that the JST-RR model achieves the best overall performance with the lowest perplexity among all models under a variety of scenarios.
In most cases, models that combine both textual reviews and overall ratings (i.e., AIR-JST, AIR-RJST, JST-RR) are superior to the models that only rely on textual reviews (i.e., JST, RJST).
It implies that the incorporation of overall ratings can effectively enhance the model prediction accuracy.
Compared to the AIR-JST and the AIR-RJST models that simply use the overall ratings as the prior parameters for the latent document-level sentiment distributions,
the proposed JST-RR model achieves better performance in capturing the intrinsic connection between review words and ratings,
leading to significant improvement in model prediction.

\subsection{Qualitative Performance Analysis} \label{case-3}
It is also important to examine the effectiveness of the proposed model in the extraction of topics and sentiments from the data.
As the estimated word distribution is conditioned on both sentiment and topic assignments, one can refer to the most frequent words (or top words) under each combination of sentiment-topic assignments for understanding the extracted topics with sentiment orientations.
Table~\ref{tab:2} shows the top positive and negative words under five example topics extracted from the \textit{Dell} dataset.
Each topic covers a specific quality aspect of Dell products as well as related services such as battery (topic 1), memory \& speed (topic 2), shipping \& return (topic 3), network connections (topic 4), and peripherals (topic 5).
In terms of sentiment, it can be seen that most of the positive words and negative words under each topic carry the corresponding sentiments well.
Some of the words (e.g., ``good'', ``not\_work'') show a general tendency of customer opinions that is independent of topics, and these words tend to appear under multiple topics frequently. Some other words could bear topic-specific sentiments.
For example, words such as ``crash'', ``burn'' are frequently used for conveying negative sentiment with respect to the topic of memory and speed (topic 2).
        \begin{sidewaystable}[htbp]
	    	\centering
	    	\caption{Example of topics under different sentiment labels in \textit{Dell} dataset extracted by JST-RR model. }
	    	\label{tab:2}
	    	\renewcommand\arraystretch{1}
	    	\setlength{\tabcolsep}{1pt}{
	    		\begin{tabular}{|cc|cc|cc|cc|cc|}     	
	    			\hline
	    			\multicolumn{2}{|c|}{Topic 1} & \multicolumn{2}{|c|}{Topic 2} & \multicolumn{2}{|c|}{Topic 3} & \multicolumn{2}{|c|}{Topic 4} & \multicolumn{2}{|c|}{Topic 5}\\
	    			
	    			Positive & Negative & Positive & Negative & Positive & Negative & Positive & Negative & Positive & Negative \\
	    			\hline
	    			batteri & problem & gb & drive & great & amazon & use & problem & screen & screen \\
	    			power & hour & ram & hard & ship & return & internet & connect & keyboard & keyboard \\
	    			use & year & processor & dvd & arriv & receiv & work & internet & mous & use \\
	    			good & replac & drive & cd & fast & sent & web & wireless & use & touch \\
	    			life & power & memori & not\_work & amazon & seller & offic & issu & like & key \\
	    			hour & month & hard & comput & well & order & wireless & time & feel & pad \\
	    			still & bought & core & replac & seller & product & great & wifi & nice & button \\
	    			year & batteri & card & littl & time & back & home & tri & good & mous \\
	    			get & motherboard & ghz & old & good & ship & connect & card & speaker & type \\
	    			like & first & intel & disk & order & refund & program & driver & featur & finger \\
	    			time & issu & speed & burn & came & disappoint & surf & fix & key & annoy \\
	    			last & time & hd & usb & product & item & run & seem & pro & plastic \\
	    			work & turn & graphic & bad & receiv & day & open & slow & light & start \\
	    			charg & last & cpu & instal & recommend & contact & fast & not\_work & tablet & issu \\
	    			thing & day & dual & fail & condit & send & easi & return & display & back \\
	    			great & charg & dvd & player & got & box & basic & minut & touch & click \\
	    			overal & repair & slot & port & packag & refurbish & like & network & qualiti & touchpad \\
	    			need & ago & perform & disc & box & arriv & load & open & model & open \\
	    			review & board & ssd & crash & happi & get & download & updat & easi & left \\
	    			well & start & pentium & ram & thank & week & love & old & inspiron & fan \\
	    			\hline
	    		\end{tabular}
	    	}
	    \end{sidewaystable}

\begin{figure}[htbp]
	    	\centering
	    	\subfigure[$\boldsymbol{\mu}^{Dell}$]{
	    		\label{fig4-a}
	    		\begin{minipage}{0.47\linewidth}
	    			\centering
	    			\includegraphics[scale=0.35]{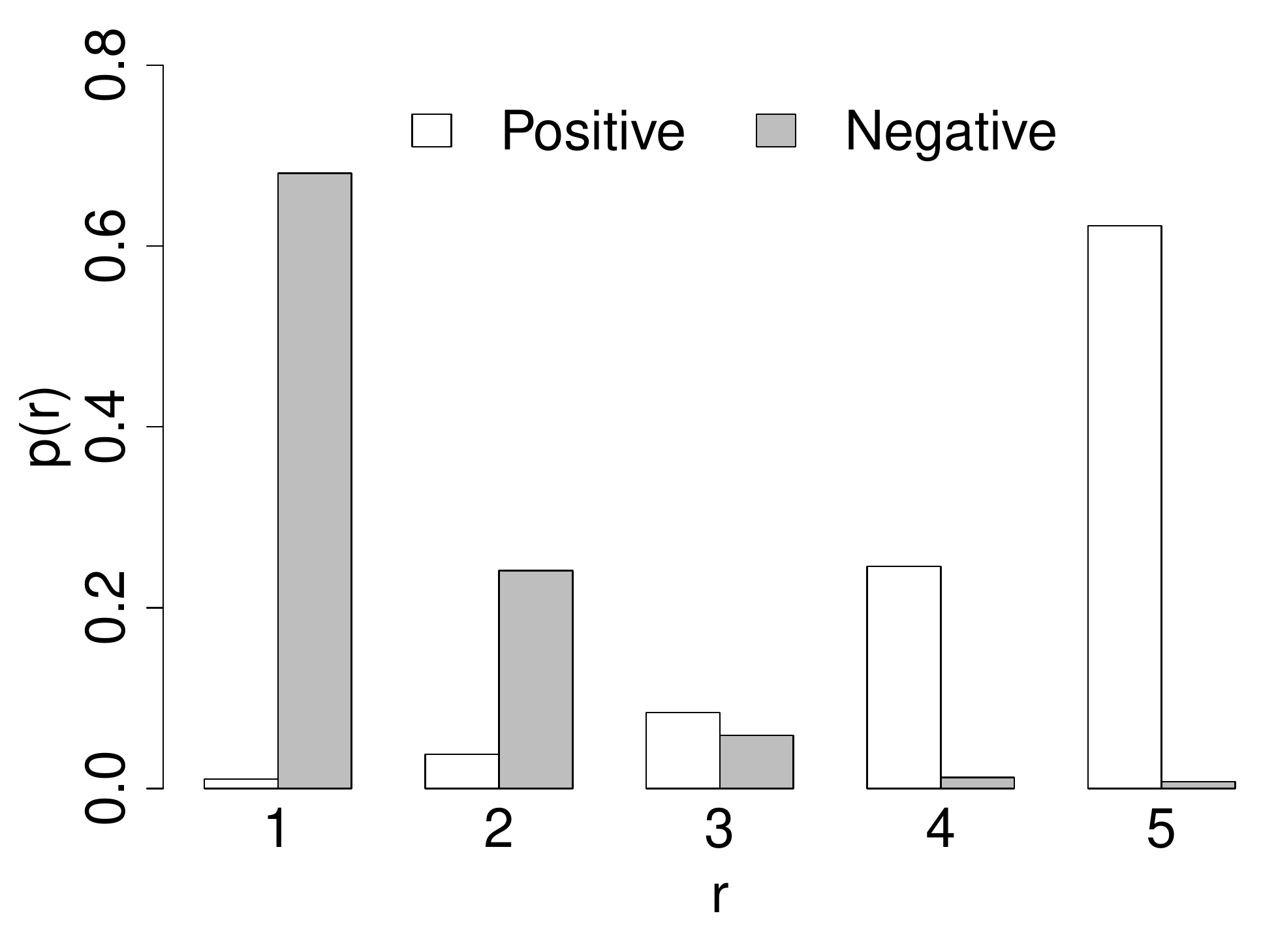}
	    		\end{minipage}
	    	}
    	    \subfigure[$\boldsymbol{\mu}^{Lenovo}$]{
    	    	\label{fig4-b}
    	    	\begin{minipage}{0.47\linewidth}
    	    		\centering
    	    		\includegraphics[scale=0.35]{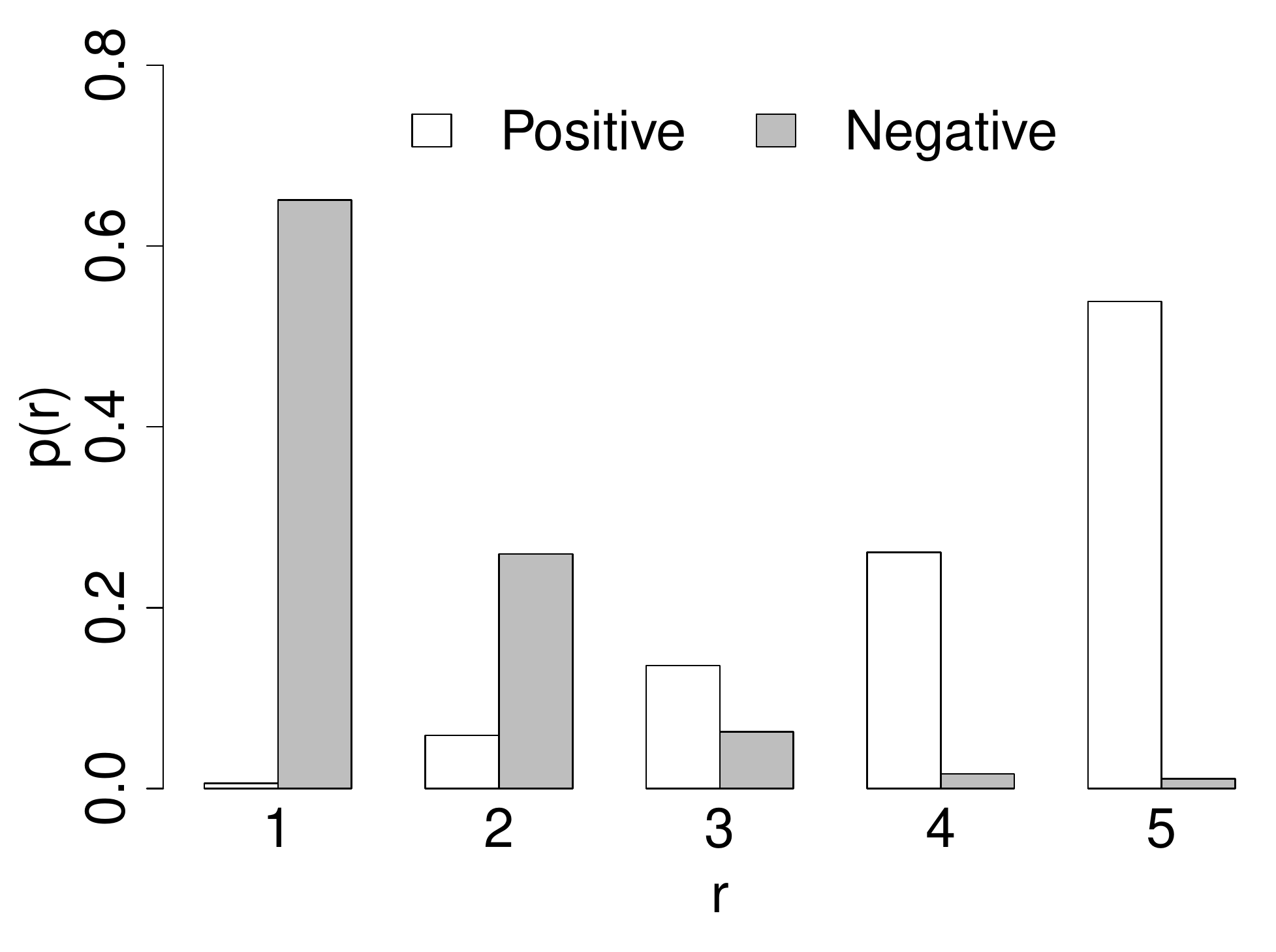}
    	    	\end{minipage}
    	    }    	
            \subfigure[$\boldsymbol{\mu}^{HP}$]{
            	\label{fig4-c}
            	\begin{minipage}{0.47\linewidth}
            		\centering
            		\includegraphics[scale=0.35]{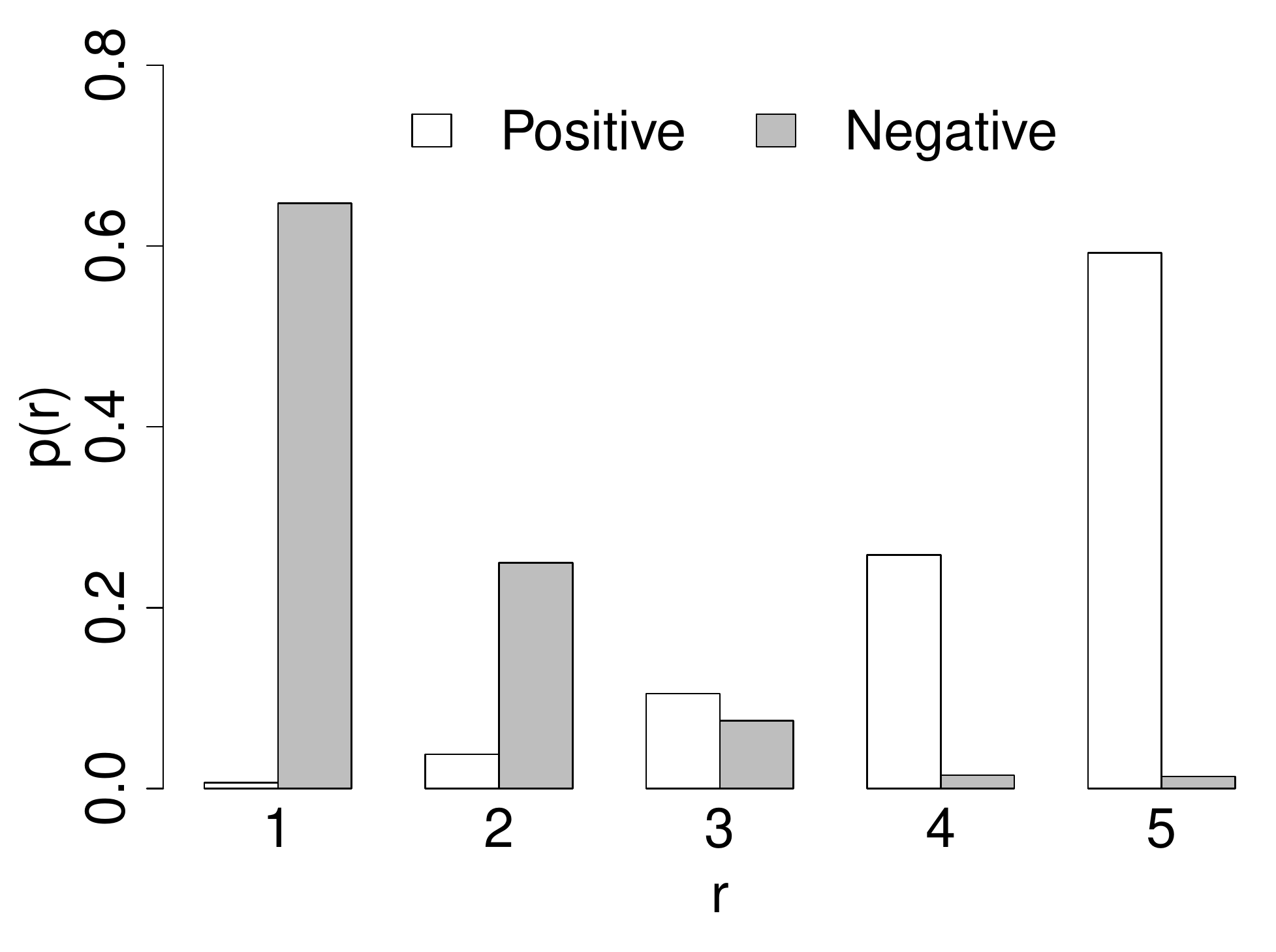}
            	\end{minipage}
            }
	    	\subfigure[$\boldsymbol{\mu}^{diff}$]{
	    		\label{fig4-d}
	    		\begin{minipage}{0.47\linewidth}
	    			\centering
	    			\includegraphics[scale=0.35]{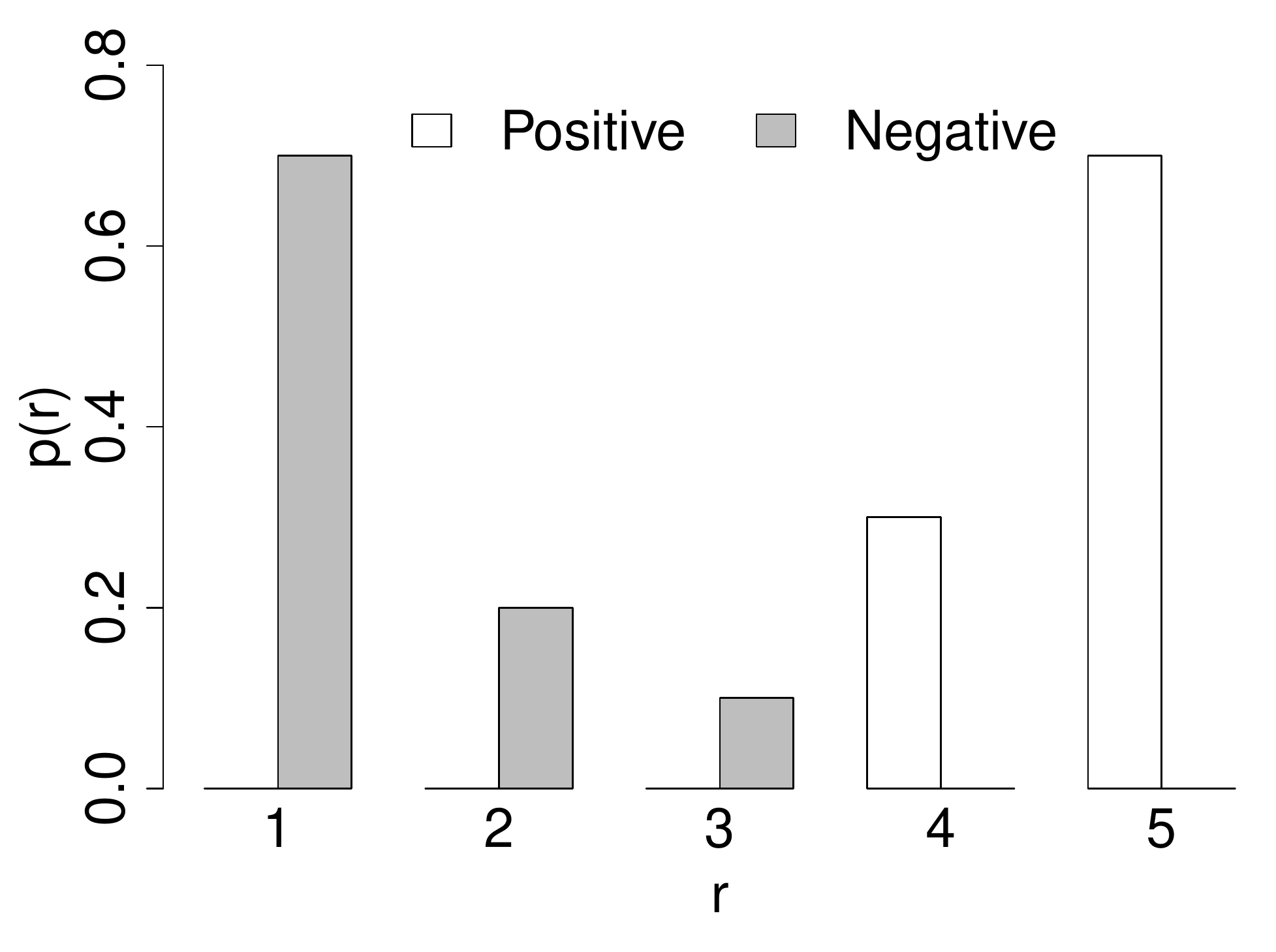}
	    		\end{minipage}
	    	}
    	    \subfigure[$\boldsymbol{\mu}^{unif}$]{
    	    	\label{fig4-e}
    	    	\begin{minipage}{0.47\linewidth}
    	    		\centering
    	    		\includegraphics[scale=0.35]{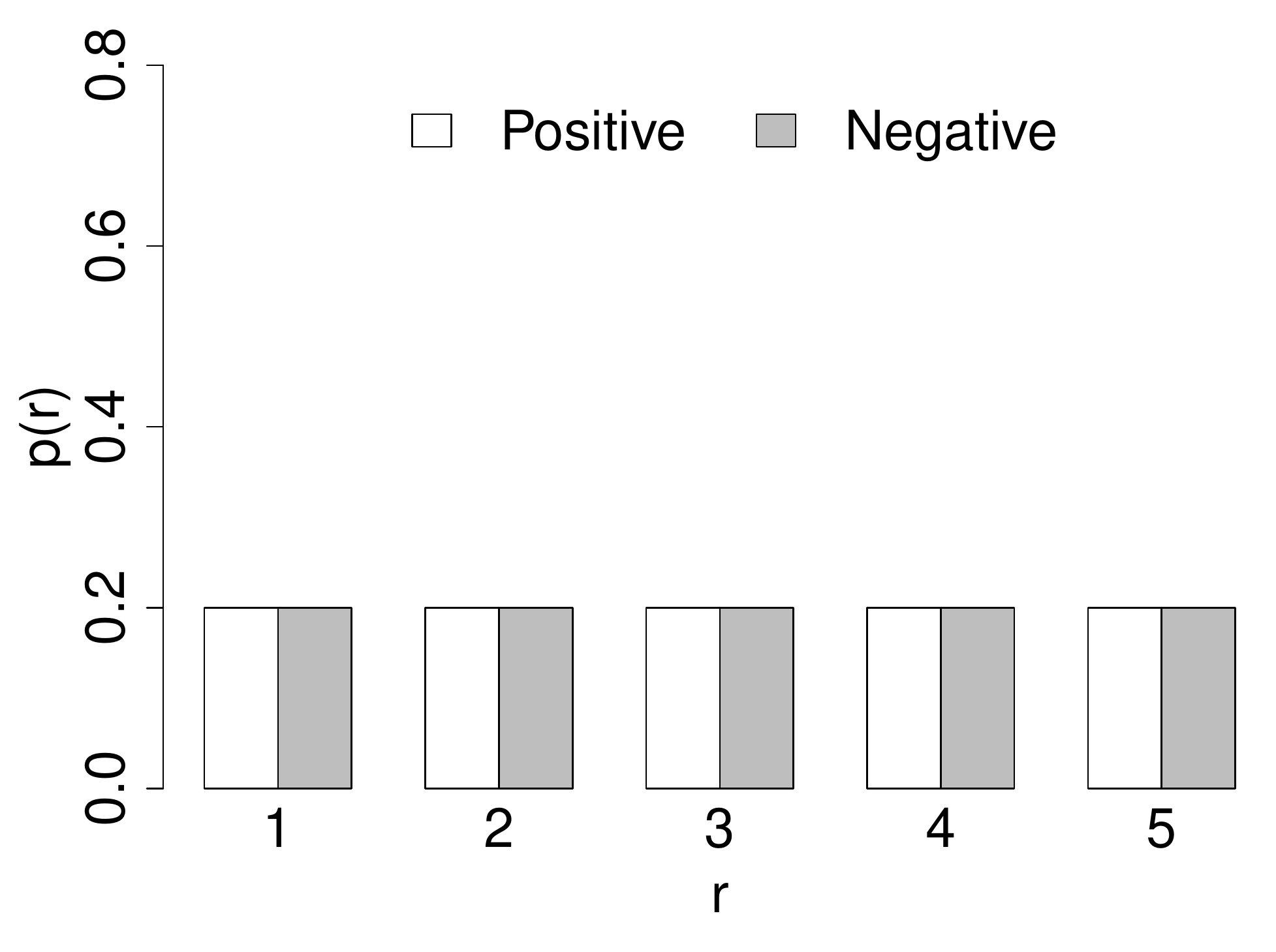}
    	    	\end{minipage}
    	    }
	    	\caption{Distributions over ratings under positive and negative sentiments.}
	    	\label{fig4}
	    \end{figure}

Moreover, a more general sentiment detection can be examined by the estimated rating distribution.
For example, Figures~\ref{fig4-a}, \ref{fig4-b}, \ref{fig4-c} show the estimated rating distribution parameter $\boldsymbol{\hat{\mu}}$ of the three experimental datasets under different sentiment labels with the topic number $K=5$.
It is seen that the positive and negative sentiments are obviously distinguished by their distributions over five rating scores.
Such an observation is validated by the results that a positive sentiment tends to produce higher ratings than the negative one, showing consistency with human expectations.
Overall, the results above demonstrate that the proposed JST-RR model enables an informative and coherent extraction of both topics and sentiments from the data.

\section{Simulation} \label{simulation}
This section conducts several simulation studies to examine the model performance in predicting the document-level sentiment distributions under various scenarios.

\subsection{Simulated documents}
We simulate review documents that are composed of words and ratings with known parameters based on the generative process in Procedure~\ref{alg1}.
Specifically, each simulated document is represented by a random joint sentiment-topic distribution $P(l,z)=\pi_l \theta_{l,z}$ that quantifies how likely the current document is linked to each sentiment and topic label.
We let the number of topics $K = 5$ and the number of sentiments $S=2$.
For each review document, we test with the number of ratings $M \in \{1,2,3,4,5,7,10\}$ and the number of words $N \in \{10M,20M,30M\}$ for each value of $M$.
Given the sentiment-topic mixtures sampled from $P(l,z)$, a simulated document is generated by sampling words and ratings from the empirical word distribution $\text{Multinomial}(\bm{\varphi})$ and rating distribution $\text{Multinomial}(\bm{\mu})$, respectively.
Without loss of generality, we use the empirical word distribution estimated from the real-world \textit{Dell} dataset in Section~\ref{case} for generating the words in simulated documents.
In addition, all the ratings are sampled from the empirical rating distribution with parameter $\boldsymbol{\mu}^{Dell}$ in Figure~\ref{fig4-a} conditioned on their sentiment assignments.

        %

Accordingly, the rating distribution provides occurrence rules among the observed ratings.
For example, based on the rating distribution with parameter $\boldsymbol{\mu}^{Dell}$ in Figure~\ref{fig4-a},
a positive sentiment is more likely to stimulate a higher rating, while a negative sentiment leads to a lower one.
Note that the rating distribution varies with the studied dataset,
and the simulation data generated with various rating distributions would lead to different results.
For conducting a general comparison, our simulation additionally explores two distant cases of rating distributions with the parameters shown in Figure~\ref{fig4-d} and Figure~\ref{fig4-e}.
Figure~\ref{fig4-d} represents an extreme case ($\boldsymbol{\mu}^{diff}$) that ratings under two sentiment classifications are totally differentiated.
In contrast, Figure~\ref{fig4-e} represents the opposite case ($\boldsymbol{\mu}^{unif}$) that ratings under two sentiment classifications are totally mixed.
In practice, the distributions over ratings would range between $\boldsymbol{\mu}^{diff}$ and $\boldsymbol{\mu}^{unif}$.

Based on Shannon's concept of information theory, the information gain (IG) on the prediction of sentiments $l\in\{1,\dots,S\}$ given specific ratings $r\in\{1,2,3,4,5\}$ is defined as
	    \begin{equation} \label{eq:simulation-1}
	    \begin{split}
	    \text{IG}(l,r)&=\text{H}(l)-\text{H}(l|r) =\sum_{r=1}^{5}P(r)\sum_{l=1}^{S}P(l|r)\log P(l|r)-\sum_{l=1}^{S}P(l)\log P(l),
	    \end{split}
	    \end{equation}
        which can be regarded as the amount of reduced randomness in predicting a sentiment given a rating. It is easy to show that the information gain in Eq.~(\ref{eq:simulation-1}) is maximized, namely $\text{H}(l|r)=0$ and $\text{IG}(l,r)=\text{H}(l)$, in the case of $\boldsymbol{\mu}^{diff}$ (Figure~\ref{fig4-d}) where the sentiment prediction is 100\% confirmed under each possible rating score. In contrast, it is minimized, namely $\text{IG}(l,r)=0$, at the uniform distribution of $\boldsymbol{\mu}^{unif}$ (Figure~\ref{fig4-e}).

\subsection{Comparison results}
Note that the incorporation of overall ratings mainly makes a difference in the estimation of document-level sentiments.
Thus we focus on the accuracy of estimating the sentiment distribution parameter $\boldsymbol{\pi}$ with the proposed Gibbs sampling procedure under different model implementations. Specifically,
the Kullback Leibler (KL) Divergence \citep{kullback1997information} is used to evaluate the performance measure of sentiment prediction as
	    \begin{equation}
	    D_{\text{KL}}(\hat{\boldsymbol{\pi}},\boldsymbol{\pi})=\sum_{l}\hat{\pi}_l\log\frac{\hat{\pi}_l}{\pi_l}.
	    \end{equation}
It measures the distance between the predicted sentiment distribution $\hat{\boldsymbol{\pi}}$ from different models and the target sentiment distribution $\boldsymbol{\pi}$ (ground-truth).
For a general comparison, we consider the following four models:
	    \begin{itemize}
	    	\item JST-RR($\mu^{diff}$): The JST-RR model applied to simulated documents generated with rating distribution parameter $\boldsymbol{\mu}^{diff}$.
	    	\item JST-RR($\mu^{unif}$): The JST-RR model applied to simulated documents generated with rating distribution parameter $\boldsymbol{\mu}^{unif}$.
	    	\item JST-RR($\mu^{Dell}$): The JST-RR model applied to simulated documents generated with rating distribution parameter $\boldsymbol{\mu}^{Dell}$.
	    	\item JST: The JST model only applied to the textual (word) part of simulated documents (baseline).
	    \end{itemize}
All the models above are implemented in the same condition.
Note that we have not included the implementations of other alternative models (e.g., RJST, AIR-JST, and AIR-RJST)
since the simulated data here are based on the generative process of the JST-RR/JST model.
The tuning parameter $\sigma$ is chosen by a 10-fold cross validation on a validation set of simulated documents.
	
Table~\ref{tab:3} reports the average KL Divergence under different models, where each average value of KL Divergence is computed based on $D=1,000$ samples of documents, and the standard deviations are shown in brackets.
In general, when $N$ and $M$ (the number of words and the number of ratings) are increased in a document, the document-level sentiment parameters are estimated with higher accuracy.
Based on results, the proposed JST-RR model with $\boldsymbol{\mu}^{diff}$ appear to achieve the best performance among all scenarios.
It indicates that the incorporation of overall ratings in case of a differentiated sentiment-rating distribution is helpful for the sentiment prediction.
Note that the JST-RR model with $\boldsymbol{\mu}^{unif}$ is equivalent to the baseline model of JST (i.e., $\sigma=0$) since the ratings in this case would not contribute to the sentiment prediction.
Generally, the improvements of the JST-RR model compared to the baseline of JST can be explained by the incorporation of  ratings.
When the ratings bring larger information gain on the sentiment prediction as defined in Eq.~(\ref{eq:simulation-1}) as in the case of $\boldsymbol{\mu}^{diff}$,
the improvements would be more significant.
In contrast, when the ratings are non-informative as in the case of $\boldsymbol{\mu}^{unif}$, the improvements are marginal.
	    \begin{table}[htbp]
	    	\centering
	    	\caption{Results of KL Divergence between the predicted and the target sentiment distributions. $M$ and $N$ represent the number of ratings and words in each document.}
	    	\label{tab:3}
	    	\renewcommand\arraystretch{1.5}
	    	\begin{tabular}{ccccc}
	    	    \hline
	    	    \multicolumn{5}{c}{N/M=10} \\
	    	    M & N & JST/JST-RR($\mu^{unif}$) & JST-RR($\mu^{Dell}$) & JST-RR($\mu^{diff}$) \\
	    		\hline
	    		1 & 10 & 0.1491(0.0059) & 0.1324(0.0052) & \textbf{0.1190(0.0047)} \\
                2 & 20 & 0.0734(0.0036) & 0.0565(0.0025) & \textbf{0.0541(0.0021)} \\
                3 & 30 & 0.0500(0.0022) & 0.0386(0.0017) & \textbf{0.0333(0.0014)} \\
                4 & 40 & 0.0375(0.0018) & 0.0298(0.0013) & \textbf{0.0225(0.0011)} \\
                5 & 50 & 0.0291(0.0013) & 0.0199(0.0008) & \textbf{0.0195(0.0008)} \\
                7 & 70 & 0.0190(0.0008) & 0.0146(0.0007) & \textbf{0.0127(0.0005)} \\
                10 & 100 & 0.0125(0.0005) & 0.0096(0.0004) & \textbf{0.0094(0.0004)} \\
                \hline
                \multicolumn{5}{c}{N/M=20} \\
	    	    M & N & JST/JST-RR($\mu^{unif}$) & JST-RR($\mu^{Dell}$) & JST-RR($\mu^{diff}$) \\
	    		\hline
	    		1 & 20 & 0.0688(0.0028) & 0.0627(0.0026) & \textbf{0.0574(0.0029)} \\
                2 & 40 & 0.0320(0.0014) & 0.0283(0.0014) & \textbf{0.0240(0.0011)} \\
                3 & 60 & 0.0216(0.0010) & 0.0191(0.0008) & \textbf{0.0172(0.0008)} \\
                4 & 80 & 0.0155(0.0007) & 0.0136(0.0006) & \textbf{0.0135(0.0006)} \\
                5 & 100 & 0.0128(0.0006) & 0.0104(0.0005) & \textbf{0.0098(0.0005)} \\
                7 & 140 & 0.0073(0.0003) & 0.0066(0.0003) & \textbf{0.0058(0.0003)} \\
                10 & 200 & 0.0054(0.0002) & 0.0046(0.0002) & \textbf{0.0046(0.0002)} \\
	    		\hline
	    		\multicolumn{5}{c}{N/M=30} \\
	    	    M & N & JST/JST-RR($\mu^{unif}$) & JST-RR($\mu^{Dell}$) & JST-RR($\mu^{diff}$) \\
	    		\hline
	    		1 & 30 & 0.0515(0.0023) & 0.0461(0.0021) & \textbf{0.0440(0.0021)} \\
                2 & 60 & 0.0227(0.0009) & 0.0194(0.0009) & \textbf{0.0183(0.0008)} \\
                3 & 90 & 0.0140(0.0006) & 0.0121(0.0005) & \textbf{0.0120(0.0005)} \\
                4 & 120 & 0.0095(0.0005) & 0.0087(0.0004) & \textbf{0.0076(0.0004)} \\
                5 & 150 & 0.0077(0.0003) & 0.0068(0.0003) & \textbf{0.0065(0.0003)} \\
                7 & 210 & 0.0055(0.0002) & 0.0046(0.0002) & \textbf{0.0041(0.0002)} \\
                10 & 300 & 0.0036(0.0002) & 0.0031(0.0001) & \textbf{0.0029(0.0001)} \\
	    		\hline
	    	\end{tabular}
	    \end{table}

\begin{figure}[htbp]
	    	\centering
	    	\subfigure[$N/M=10$]{
	    		\label{fig5-a}
	    		\begin{minipage}{0.47\linewidth}
	    			\centering
	    			\includegraphics[scale=0.45]{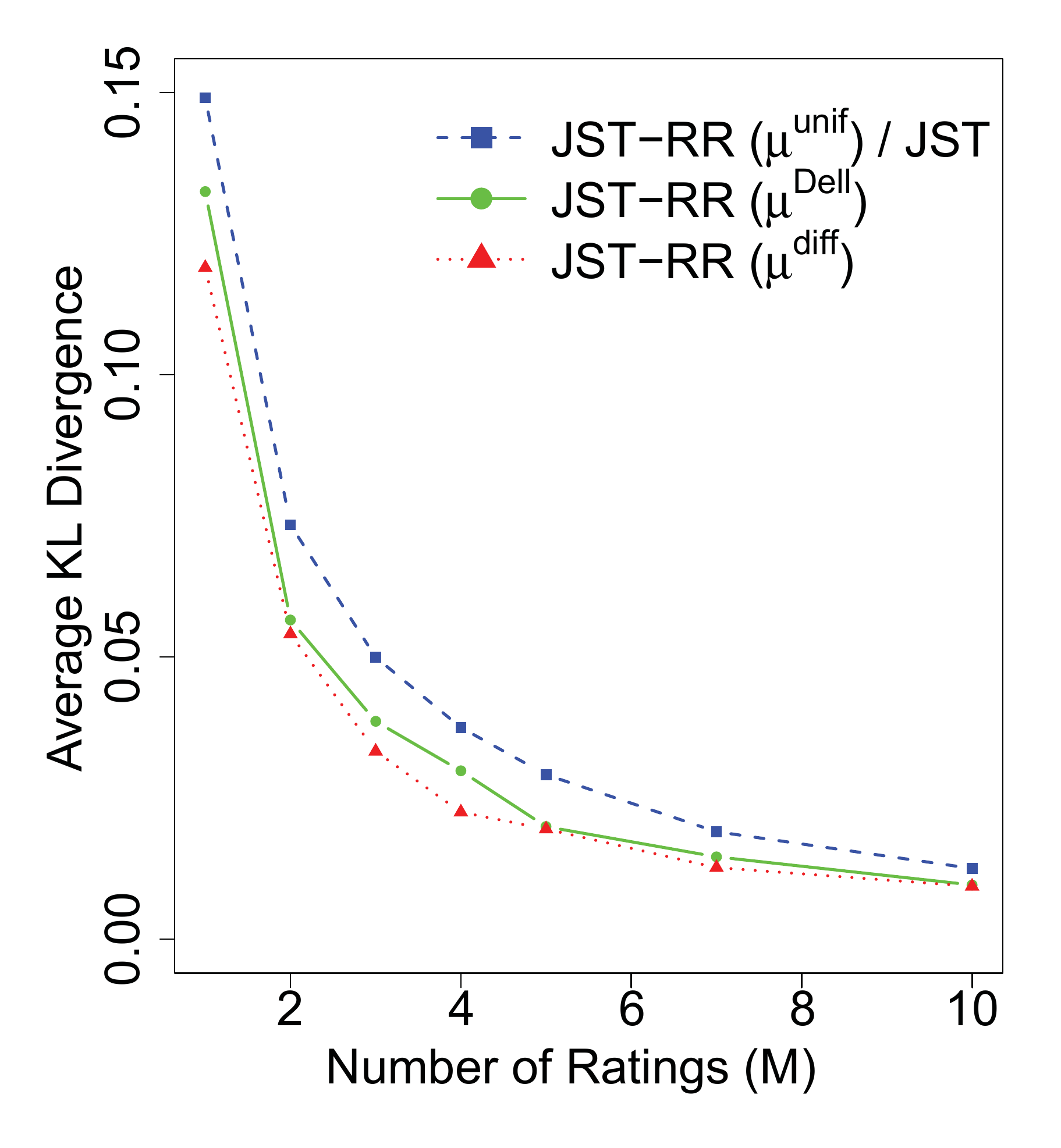}
	    		\end{minipage}
	    	}     	
	    	\subfigure[$N/M=20$]{
	    		\label{fig5-b}
	    		\begin{minipage}{0.47\linewidth}
	    			\centering
	    			\includegraphics[scale=0.45]{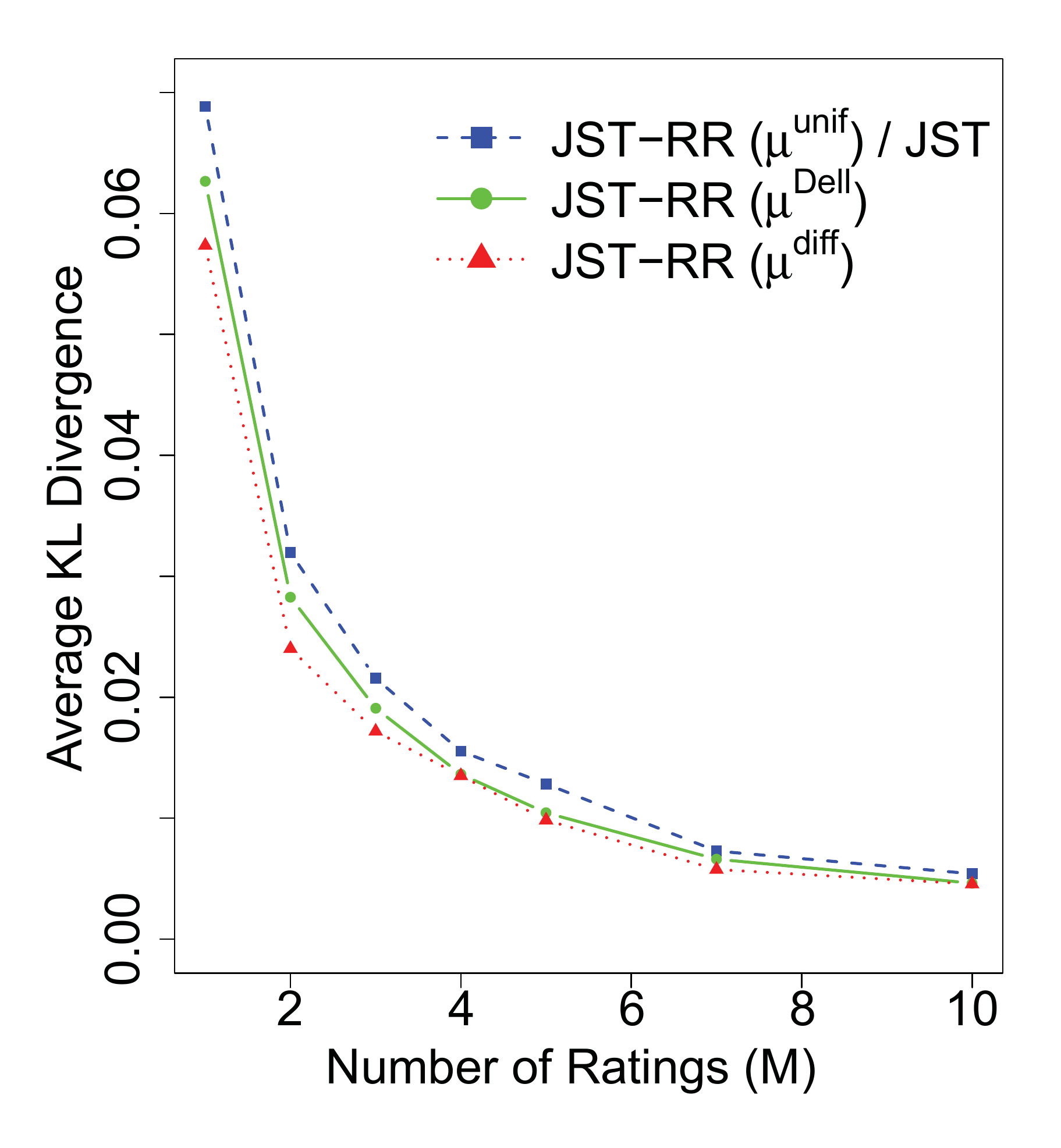}
	    		\end{minipage}
	    	}
	    	\subfigure[$N/M=30$]{
	    		\label{fig5-c}
	    		\begin{minipage}{\linewidth}
	    			\centering
	    			\includegraphics[scale=0.45]{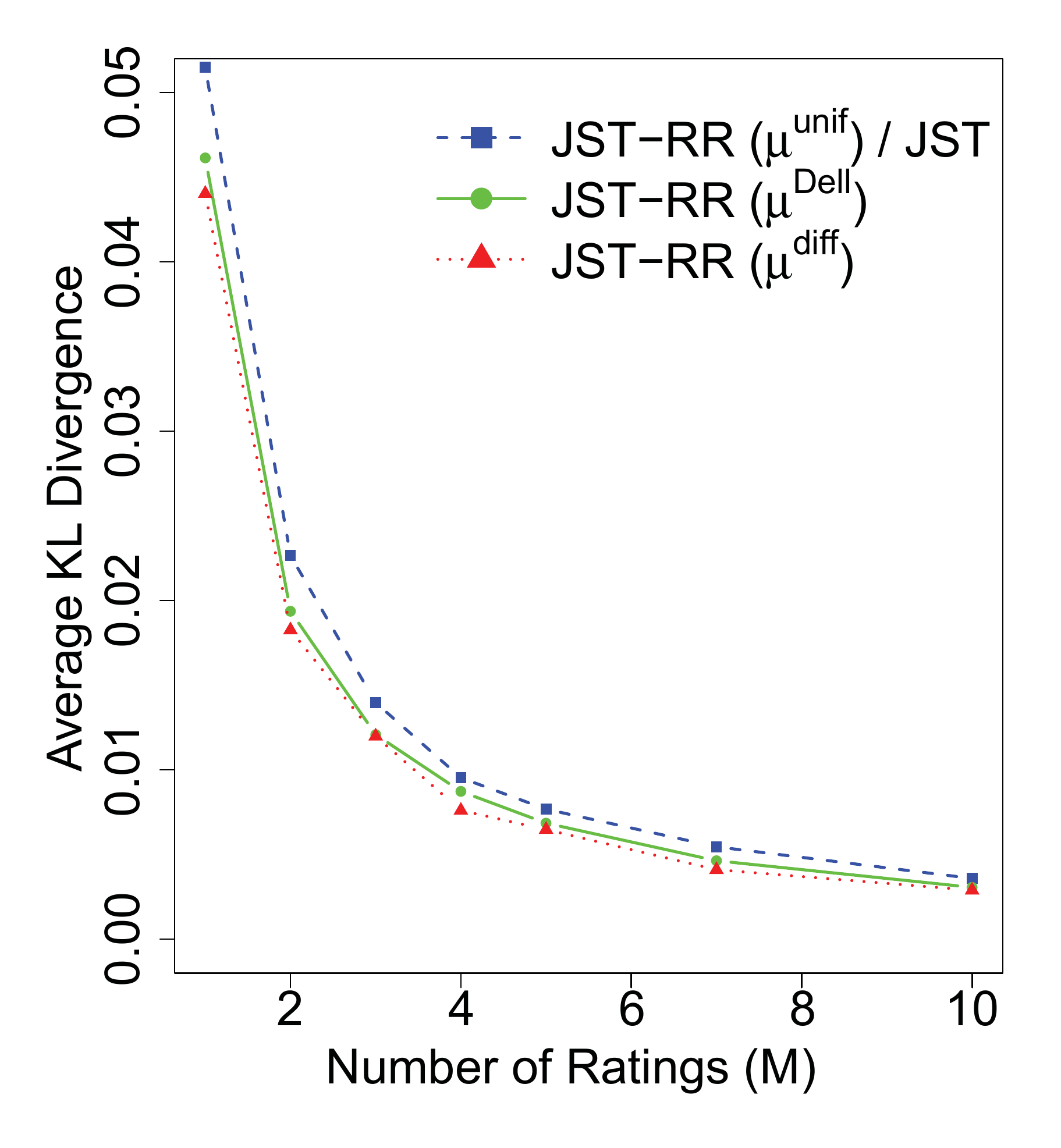}
	    		\end{minipage}
	    	}
	    	\caption{Average KL Divergence between the predicted sentiment distribution and the ground truth under different word-rating ratios ($N/M=10,20,30$).}
	    	\label{fig5}
	    \end{figure}

We also plot the results of KL Divergence in Figure~\ref{fig5} for a graphical visualization.
It is clearly seen that the improvements in sentiment prediction become smaller with an increasing word-rating ratio $N/M$ in review documents.
For example, the JST-RR model under the word-rating ratio $N/M=10$ has a significant advantage over the JST model (Figure~\ref{fig5-a}).
While the advantage is reduced with the word-rating ratio $N/M=30$ (Figure~\ref{fig5-c}).
It shows that the improvement from complementary ratings in the JST-RR model could be marginal when there are a sufficient amount of words for the document sentiment prediction.
In a short summary, the proposed JST-RR model has the advantage for short reviews with insufficient words (or a low word-rating ratio).
	
\section{Discussion} \label{conclusion}

In this work, we propose a joint sentiment-topic model to properly accommodate ratings and review texts.
The proposed model characterizes the intrinsic connection between review texts and ratings, leading to accurate prediction on review sentiments and topics.
An efficient Gibbs sampling procedure is developed to make inference for the model parameters.
Through the case study on the Amazon datasets, it appears that the proposed JST-RR model can enable an effective identification of latent topics and sentiments in reviews.
It is noted that the proposed JST-RR model brings higher improvements in sentiment prediction with a more informative rating distribution and a decreasing word-rating ratio in review documents. 

Note that the proposed model is weakly supervised with the only supervision from a domain-independent sentiment lexicon.
It can be adapted to other applications easily, such as process monitoring of online services \citep{liang2020ratings} and detection of fake news in social media.
Moreover, one can consider the ratings on some pre-specified topics, namely, aspect ratings.
In such situations, it is interesting to extend the proposed method to the case where aspect ratings are available, where the topic-sentiment correlation needs to be constructed appropriately by incorporating aspect ratings with review texts.
The current proposed method is mainly based on data from one platform, i.e., the reviews and ratings from Amazon.
Another direction for future research is to incorporate the platform information of reviews into the proposed method
such that it can integrate the reviews and ratings of the same or similar products from multiple platforms.

\section*{Appendix}
	    \setcounter{equation}{0}
	    \renewcommand{\theequation}{A.\arabic{equation}}
	    The first term in Eq.~(\ref{eq:JSTRR-1}) can be derived by integrating out the document-level sentiment distribution parameter $\bm\pi_i$ as
	    \begin{equation}
	    \label{A-1}
	    P(l^w_{ij}=l|\bm{l}^w_{-ij},\bm{l}^r)=\int_{\boldsymbol{\pi}_i} P(l^w_{ij}=l|\bm{\pi}_i)P(\bm{\pi}_i|\bm{l}^w_{-ij},\bm{l}^r)\,\mathrm{d}\bm{\pi}_i,
	    \end{equation}
	    where the second term is derived as
	    \begin{equation}
	    P(\bm{\pi}_i|\bm{l}^w_{-ij},\bm{l}^r)\propto P(\bm{l}^w_{-ij},\bm{l}^r|\bm{\pi}_i)P(\bm{\pi}_i).
	    \end{equation}
	    Since $P(\bm{\pi}_i)=\text{Dirichlet}(\bm \pi_i|\bm{\gamma})$ is conjugate to the multinomial sentiment probability:
	    \begin{equation}
	    \begin{split}
	    P(\bm{l}^w_{-ij},\bm{l}^r|\bm{\pi}_i)&=\prod_{k=1,k\neq j}^{N_i}\text{Multinomial}(l^w_{ik}|\bm \pi_i)\prod_{k=1}^{M_i}\text{Multinomial}(l^r_{ik}|\bm \pi_i)\\
	    &=\prod_{l=1}^{S}(\pi_{i,l})^{N_{i,l}^{-ij}+M_{i,l}},
	    \end{split}	
	    \end{equation}
	    the posterior is also a Dirichlet distribution: $P(\bm{\pi}_i|\bm{l}^w_{-ij},\bm{l}^r)=\text{Dirichlet}(\bm \pi_i|\bm n_i^{-ij}+\bm m_i+\bm{\gamma})$, where $\bm n_i$ and $\bm m_i$ are the counting number of words and ratings in the document $d_i$ that are associated with each sentiment label:
	    \begin{equation}
	    \bm n_i=(N_{i,1},N_{i,2},\dots,N_{i,S}),\quad \bm m_i=(M_{i,1},M_{i,2},\dots,M_{i,S}).
	    \end{equation}
	
	    \begin{remark}
	    {\it When one would like to incorporate a weighting mechanism between the number of words and ratings in sentiment estimation,
	    we can consider a weighted likelihood as
	    \begin{equation}
	    \begin{split}
	    P(\bm{l}^w_{-ij},\bm{l}^r|\bm{\pi}_i)&=\prod_{k=1,k\neq j}^{N_i}{\rm Multinomial}(l^w_{ik}|\bm \pi_i)
	    \left [\prod_{k=1}^{M_i} {\rm Multinomial}(l^r_{ik}|\bm \pi_i) \right]^{\sigma} \\
	    &=\prod_{l=1}^{S}(\pi_{i,l})^{N_{i,l}^{-ij}+\sigma M_{i,l}}.
	    \end{split}	
	    \end{equation}
        Then the posterior is still a Dirichlet distribution: $P(\bm{\pi}_i|\bm{l}^w_{-ij},\bm{l}^r)={\rm Dirichlet}(\bm \pi_i|\bm n_i^{-ij}+ \sigma \bm m_i+\bm{\gamma})$.}
	    \end{remark}
	
	    The posterior predictive distribution of Eq.~(\ref{A-1}) can be derived as
	    \begin{equation}
	    \begin{aligned}
	    P(l^w_{ij}=l|\bm{l}^w_{-ij},\bm{l}^r)&=\int_{\boldsymbol{\pi}_i} P(l^w_{ij}=l|\bm{\pi}_i)P(\bm{\pi}_i|\bm{l}^w_{-ij},\bm{l}^r)\,\mathrm{d}\bm{\pi}_i\\
	    &=\int_{\boldsymbol{\pi}_i} P(l^w_{ij}=l|\bm{\pi}_i)\text{Dirichlet}(\bm \pi_i|\bm n_i^{-ij}+\bm m_i+\bm{\gamma})\,\mathrm{d}\bm{\pi}_i\\
	    &=\int_{\boldsymbol{\pi}_i} \pi_{i,l}\cdot\text{Dirichlet}(\bm \pi_i|\bm n_i^{-ij}+\bm m_i+\bm{\gamma})\,\mathrm{d}\bm{\pi}_i\\
	    &=E(\pi_{i,l}|\text{Dirichlet}(\bm \pi_i|\bm n_i^{-ij}+\bm m_i+\bm{\gamma})),
	    \end{aligned}	
	    \end{equation}
	    which is the expected value of $\text{Dirichlet}(\bm \pi_i|\bm n_i^{-ij}+\bm m_i+\bm{\gamma})$ on the sentiment dimension of $l$. According to the expected value of Dirichlet distribution in the Dirichlet-multinomial conjugate framework, we can obtain the final derivation of Eq.~(\ref{A-1}) as
	    \begin{equation}
	    \begin{aligned}
	    P(l^w_{ij}=l|\bm{l}^w_{-ij},\bm{l}^r)&=E(\pi_{i,l}|\text{Dirichlet}(\bm \pi_i|\bm n_i^{-ij}+\bm m_i+\bm{\gamma}))\\
	    &=\frac{N_{i,l}^{-ij}+M_{i,l}+\gamma_{l}}{N_i^{-ij}+M_i+\sum_{l'}\gamma_{l'}},
	    \end{aligned}	
	    \end{equation}
	    where $N_i$ and $M_i$ are the total number of words and ratings in the document $d_i$, $N_{i,l}$ and $M_{i,l}$ are the number of words and ratings associated with sentiment $l$ in the document $d_i$, and the hyperparameter $\gamma_l$ can be interpreted as the prior observation counts of the sentiment $l$ assigned with $d_i$.
	
	    Similarly, the second term in Eq.~(\ref{eq:JSTRR-1}) can be derived by integrating out the random variable $\boldsymbol{\theta}_{i,l}$ as:
	    \begin{equation}
	    \label{A-2}
	    P(z_{ij}=z|l^w_{ij}=l,\bm{l}^w_{-ij},\bm{z}_{-ij})=\int_{\bm{\theta}_{i,l}} P(z_{ij}=z|\bm{\theta}_{i,l})P(\bm{\theta}_{i,l}|\bm{l}^w_{-ij},\bm{z}_{-ij})\,\mathrm{d}\bm{\theta}_{i,l},	
	    \end{equation}
	    where the second term is derived as
	    \begin{equation}
	    P(\bm{\theta}_{i,l}|\bm{l}^w_{-ij},\bm{z}_{-ij})\propto P(\bm{z}_{-ij}|\bm{\theta}_{i,l},\bm{l}^w_{-ij})P(\bm{\theta}_{i,l}).
	    \end{equation}
	    Since $P(\bm{\theta}_{i,l})=\text{Dirichlet}(\bm{\theta}_{i,l}|\bm{\alpha}_{l})$ is conjugate to the multinomial probability $P(\bm{z}_{-ij}|\bm{\theta}_{i,l},\bm{l}^w_{-ij})$, the posterior is also a Dirichlet distribution: $P(\bm{\theta}_{i,l}|\bm{l}^w_{-ij},\bm{z}_{-ij})=\text{Dirichlet}(\bm{\theta}_{i,l}|\bm n^{-ij}_{i,l}+\bm{\alpha}_{l})$, where $\bm n_{i,l}=(N_{i,l,1},N_{i,l,2},\dots,N_{i,l,K})$. By following the same derivation, the posterior predictive distribution of Eq.~(\ref{A-2}) is
	    \begin{equation}
	    \begin{aligned}
	    P(z_{ij}=z|l^w_{ij}=l,\bm{l}^w_{-ij},\bm{z}_{-ij})&=\int_{\bm{\theta}_{i,l}} P(z_{ij}=z|\bm{\theta}_{i,l})P(\bm{\theta}_{i,l}|\bm{l}^w_{-ij},\bm{z}_{-ij})\,\mathrm{d}\bm{\theta}_{i,l}\\
	    &=\int_{\bm{\theta}_{i,l}} \theta_{i,l,z}\cdot\text{Dirichlet}(\bm{\theta}_{i,l}|\bm n^{-ij}_{i,l}+\bm{\alpha}_{l})\,\mathrm{d}\bm{\theta}_{i,l}\\
	    &=E(\theta_{i,l,z}|\text{Dirichlet}(\bm{\theta}_{i,l}|\bm n^{-ij}_{i,l}+\bm{\alpha}_{l}))\\
	    &=\frac{N^{-ij}_{i,l,z}+\alpha_{l,z}}{N^{-ij}_{i,l}+\sum_{z'}\alpha_{l,z'}},
	    \end{aligned}
	    \end{equation}
	    where $N_{i,l,z}$ is the number of words associated with the sentiment $l$ and topic $z$ in the document $d_i$, and the hyperparameter $\alpha_{l,z}$ can be interpreted as the prior observation counts of words assigned with the sentiment $l$ and topic $z$ in $d_i$.
	
	    Similarly, for the third term in Eq.~(\ref{eq:JSTRR-1}), we can obtain its derivation by integrating out the variable $\boldsymbol{\varphi}_{l,z}$ as
	    \begin{equation}
	    \label{A-3}
	    \begin{aligned}
	    P(w_{ij}=w|l^w_{ij}=l,z_{ij}=z,\bm{l}^w_{-ij},\bm{z}_{-ij},\bm{w}_{-ij})&=\int_{\bm{\varphi}_{l,z}} P(w_{ij}=w|\bm{\varphi}_{l,z})P(\bm{\varphi}_{l,z}|\bm{l}^w_{-ij},\bm{z}_{-ij},\bm{w}_{-ij})\,\mathrm{d}\bm{\varphi}_{l,z}\\
	    &=\int_{\bm{\varphi}_{l,z}} \varphi_{l,z,w}\cdot\text{Dirichlet}(\bm{\varphi}_{l,z}|\bm n^{-ij}_{l,z}+\bm{\beta}_{l,z})\,\mathrm{d}\bm{\varphi}_{l,z}\\
	    &=E(\varphi_{l,z,w}|\text{Dirichlet}(\bm{\varphi}_{l,z}|\bm n^{-ij}_{l,z}+\bm{\beta}_{l,z}))\\
	    &=\frac{N_{l,z,w}^{-ij}+\beta_{l,z,w}}{N_{l,z}^{-ij}+\sum_{w'}\beta_{l,z,w'}},
	    \end{aligned}	
	    \end{equation}
	    where $N_{l,z}$ is the number of words assigned with the sentiment label $l$ and topic label $z$ in the entire dataset, $\bm n_{l,z}=(N_{l,z,1},\dots,N_{l,z,V})$ is the number of times that each word $w\in\{1,\dots,V\}$ is associated with the sentiment label $l$ and topic label $z$ in the dataset, and the hyperparameter $\beta_{l,z,w}$ can be interpreted as the prior counts of word $w$ associated with sentiment label $l$ and topic label $z$ in the dataset.
	
	    Similarly, the second term in Eq.~(\ref{eq:JSTRR-7}) can be derived by integrating out $\bm{\mu}_{l}$ as
	    \begin{equation}
	    \label{A-4}
	    \begin{aligned}
	    P(r_{ij}=r|l^r_{ij}=l,\bm{l}^r_{-ij},\bm{r}_{-ij})&=\int_{\bm{\mu}_{l}} P(r_{ij}=r|\bm{\mu}_{l})P(\bm{\mu}_{l}|\bm{l}^r_{-ij},\bm{r}_{-ij})\,\mathrm{d}\bm{\mu}_{l}\\
	    &=\int_{\bm{\mu}_{l}} \mu_{l,r}\cdot\text{Dirichlet}(\bm{\mu}_{l}|\bm m^{-ij}_{l}+\bm{\delta}_{l})\,\mathrm{d}\bm{\mu}_{l}\\
	    &=E(\mu_{l,r}|\text{Dirichlet}(\bm{\mu}_{l}|\bm m^{-ij}_{l}+\bm{\delta}_{l}))\\
	    &=\frac{M_{l,r}^{-ij}+\delta_{l,r}}{M_{l}^{-ij}+\sum_{r'}\delta_{l,r'}},
	    \end{aligned}	
	    \end{equation}
	    where $M_{l}$ is the number of ratings associated with the sentiment $l$ in the dataset, $\bm m_l=(M_{l,1},\dots,M_{l,5})$ is the number of times that each rating $r\in\{1,2,3,4,5\}$ is associated with the sentiment label $l$ in the dataset, and the hyperparameter $\delta_{l,r}$ can be interpreted as the prior counts of rating $r$ associated with sentiment label $l$ in the dataset.
	
	\bibliography{Ref}
	

\end{document}